\documentclass{article}

\usepackage[preprint]{tmlr}

\let\TMLRAnd\AND
\let\AND\undefined

\usepackage{amsmath,amssymb,amsfonts}
\usepackage{algorithmic}
\usepackage{algorithm}
\usepackage{graphicx}
\usepackage{textcomp}
\usepackage{xcolor}
\usepackage{booktabs}
\usepackage{multirow}
\usepackage{hyperref}
\usepackage{url}
\usepackage{tikz}
\usetikzlibrary{shapes,arrows,positioning,fit,backgrounds,calc}
\usepackage{subcaption}

\let\cite\citep

\title{When Does Sparse MoE Help in Vision?\\The Role of Backbone Compute Leverage in Sparse Routing}

\author{%
  \name Libo Sun \email libo@auburn.edu \\
  \addr Department of Computer Science and Software Engineering\\ Auburn University, Auburn, AL 36849, USA
  \TMLRAnd
  \name Po-wei Harn \email harnpowei@ncu.edu.tw \\
  \addr Department of Information Management\\ National Central University, Taoyuan 320317, Taiwan
  \TMLRAnd
  \name Peixiong He \email pzh0029@auburn.edu \\
  \addr Department of Computer Science and Software Engineering\\ Auburn University
  \TMLRAnd
  \name Xiao Qin \email xqin@auburn.edu \\
  \addr Department of Computer Science and Software Engineering\\ Auburn University
}

\def\month{00}
\def\year{0000}
\def\openreview{\url{https://openreview.net/forum?id=XXXXXX}}

\begin{document}

\maketitle

\begin{abstract}
Mixture-of-Experts (MoE) networks promise favorable accuracy--compute trade-offs, yet practical vision deployments are hindered by expert collapse and limited end-to-end efficiency gains. We study when sparse top-$k$ routing with hard capacity constraints helps in vision classification, evaluated under multi-seed protocols on four benchmarks (CIFAR-10/100, Tiny-ImageNet, ImageNet-1K). We observe a \emph{compute-leverage pattern}: positive accuracy gaps require a substantial fraction $\rho$ of total FLOPs to be routed; at ImageNet scale this is necessary but not sufficient, as multi-expert routing ($k \geq 2$) is additionally required. Two controlled experiments isolate these factors. A hidden-size sweep on CIFAR-10 yields both predicted sign reversals across standard and depthwise backbones, ruling out backbone family as the active variable. An ImageNet-1K ablation that varies only top-$k$---holding architecture, initialization, and $\rho$ fixed---reverses the gap from positive to negative across all five seeds. A per-sample variant of Soft MoE that softmaxes over experts rather than the batch rescues CIFAR-100 above the dense baseline, identifying batch-axis dispatch as the dominant failure mode in per-sample CNN settings. Code and aggregate results: \url{https://github.com/libophd/sparse-moe-vision-rho}.
\end{abstract}

\section{Introduction}
\label{sec:introduction}

Mixture-of-Experts (MoE) architectures route inputs through specialized subnetworks~\cite{jacobs1991adaptive,jordan1994hierarchical}. Sparsely-gated variants---activating only a few experts per input---have scaled to thousands of experts in language~\cite{shazeer2017outrageously,fedus2022switch} and vision~\cite{riquelme2021scaling,han2024vimoe} while keeping per-token compute tractable. Despite these advances, two persistent obstacles hinder practical deployments: the shared backbone often dominates total cost, leaving little structural opportunity for sparse routing to improve end-to-end efficiency~\cite{riquelme2021scaling,du2022glam}; and routers frequently collapse onto a subset of experts, undermining specialization~\cite{nie2022dense,chi2022representation}.

These obstacles are coupled. When the classification head is a negligible fraction of total FLOPs---less than $0.1\%$ on ResNet-18---even perfect routing cannot meaningfully reduce end-to-end cost, and the absence of a clear compute payoff blunts incentives to diagnose collapse or refine routing. \emph{Depthwise separable convolutions}~\cite{chollet2017xception,howard2017mobilenets} shift the head-to-backbone FLOPs ratio toward parity ($\sim$49\% on CIFAR), opening a regime in which sparse routing has structural room to help. Combined with hard capacity constraints to prevent collapse, this design admits a controlled study of when sparse MoE delivers accuracy gains in vision classification, and what conditions are necessary for those gains to appear.

Our study makes the following contributions:
\begin{enumerate}
  \item An empirical \textbf{compute-leverage observation} for vision MoE: positive accuracy gaps require a substantial fraction $\rho$ of total FLOPs to be routed; at ImageNet scale, this is necessary but not sufficient, as multi-expert routing ($k \geq 2$) is additionally required. Depthwise separable backbones tend to raise $\rho$ toward parity.
  \item Two \textbf{controlled experiments separating the roles of routed compute and routing breadth}. A six-configuration hidden-size sweep on CIFAR-10 yields a monotonic relationship between gap and $\rho$ with both predicted sign reversals: a high-$\rho$ \emph{standard} backbone is positive while a low-$\rho$ \emph{depthwise} backbone is negative, ruling out backbone family as the active variable. A complementary $k$-ablation on ImageNet-1K (backbone MoE at fixed routed fraction and initialization) varies only the top-$k$ selection and reverses the gap by roughly $3.25\%$ across all five seeds.
  \item \textbf{Multi-dataset evaluation} on CIFAR-10/100, Tiny-ImageNet, and ImageNet-1K (Fig.~\ref{fig:scaling-trend}). Under depthwise co-optimization, gaps are statistically significant on the three smaller-scale benchmarks and at ImageNet scale only when MoE is placed within backbone convolutions; cross-dataset comparisons span different backbone widths and evaluation protocols.
  \item A \textbf{routing-stability analysis}: we adopt Switch-style hard capacity constraints to guarantee per-batch expert usage during training, and find that temperature scheduling is a complementary lever for late-stage stability (Section~\ref{subsec:hard-capacity}).
  \item A \textbf{cross-method dispatch diagnostic}: Soft MoE's batch-axis dispatch collapses in per-sample CNN settings; a per-sample variant that softmaxes over experts rather than over the batch rescues CIFAR-100 above dense but leaves Tiny-ImageNet and ImageNet-1K negative, identifying batch-axis dispatch as the dominant but not sole failure mode.
\end{enumerate}

\section{Related Work}
\label{sec:related-work}

\textbf{Mixture-of-Experts.}
MoE formulations partition the predictive task across specialists blended through a learned router~\cite{jacobs1991adaptive,jordan1994hierarchical}. Sparsely-activated variants route each token to a few experts, scaling to thousands of experts in language~\cite{shazeer2017outrageously,fedus2022switch} and vision~\cite{riquelme2021scaling,han2024vimoe}. Routing collapse---a few experts attracting most of the load---remains an active challenge~\cite{nie2022dense,chi2022representation}, and recent large-scale deployments such as Mixtral~\cite{jiang2024mixtral} demonstrate the approach in practice. Most relevant to our study, \citet{clark2022unified} derive unified scaling laws for routed models, formalizing the capacity--compute trade-off that motivates our compute-leverage observation.

\textbf{Routing Strategies.}
Soft MoE~\cite{puigcerver2024soft} replaces discrete assignments with differentiable slot combinations; Expert Choice~\cite{zhou2022expertchoice} inverts routing to let experts select tokens; ST-MoE~\cite{zoph2022stmoe} introduces router z-loss for training stability. Our work does not propose a new routing mechanism but instead compares against these alternatives under matched conditions to characterize when sparse routing helps.

\textbf{Vision MoE and Efficient Backbones.}
Vision MoE deployments face a compute imbalance: backbone layers dominate total cost, leaving routing little structural opportunity to help~\cite{riquelme2021scaling}. \citet{liu2024routers} report that router design matters more at smaller scales, and \citet{videau2024sweetspot} find sparse routing most effective for small- to mid-sized vision models; both observations are consistent with the FLOP-leverage pattern we identify. Depthwise separable convolutions~\cite{chollet2017xception,howard2017mobilenets,sandler2018mobilenetv2} substantially reduce convolutional cost. We leverage this property not as a stand-alone efficiency contribution but as a way to raise the routed FLOP fraction $\rho$ on which the compute-leverage condition depends.

\textbf{Dynamic and Conditional Computation.}
Dynamic networks adapt computation per input via early exit, block skipping, or token pruning~\cite{bengio2013estimating,han2022dynamic,wang2018skipnet,rao2021dynamicvit}. Mixture-of-Depths~\cite{raposo2024mod} extends sparse routing along the depth axis. Fused MoE kernels~\cite{gale2023megablocks,hwang2023tutel} address dispatch overhead at deployment, an issue we return to in our latency analysis (Section~\ref{subsec:flops-decomp}).

\textbf{Hyperparameter Search.}
Evolutionary methods~\cite{rechenberg1973evolutionsstrategie,real2019regularized} offer a lightweight alternative to grid or Bayesian search~\cite{li2017hyperband}. We use a narrow four-dimensional evolutionary search as a tuning protocol on the development set rather than a NAS contribution.

\section{Methodology}
\label{sec:methodology}

\begin{figure*}[t]
\centering
\resizebox{0.85\textwidth}{!}{%
\begin{tikzpicture}[
    node distance=0.6cm and 1.0cm,
    inputbox/.style={rectangle, draw=black!70, line width=0.6pt, fill=gray!8, rounded corners=3pt, minimum height=0.8cm, font=\normalsize, align=center},
    convblock/.style={rectangle, draw=black!70, line width=0.6pt, fill=blue!12, rounded corners=3pt, minimum height=0.75cm, text width=2.2cm, font=\small, align=center},
    routerbox/.style={rectangle, draw=teal!80, line width=0.8pt, fill=teal!8, rounded corners=3pt, minimum height=1.0cm, font=\small, align=center},
    expertbox/.style={rectangle, draw=orange!70, line width=0.6pt, fill=orange!8, rounded corners=2pt, minimum width=1.2cm, minimum height=0.85cm, font=\scriptsize, align=center},
    expertsel/.style={rectangle, draw=orange!90, line width=1.2pt, fill=orange!25, rounded corners=2pt, minimum width=1.2cm, minimum height=0.85cm, font=\scriptsize, align=center},
    utilbox/.style={rectangle, draw=red!60, line width=0.6pt, fill=red!6, rounded corners=3pt, minimum height=0.7cm, font=\small, align=center},
    gabox/.style={rectangle, draw=violet!70, line width=0.8pt, fill=violet!8, rounded corners=3pt, minimum height=0.7cm, font=\small, align=center},
    outbox/.style={rectangle, draw=black!70, line width=0.6pt, fill=green!10, rounded corners=3pt, minimum height=0.8cm, font=\normalsize, align=center},
    arr/.style={->,>=stealth, line width=0.5pt, black!70},
    thickarr/.style={->,>=stealth, line width=0.8pt, black!80},
    dasharr/.style={->,>=stealth, line width=0.5pt, dashed},
    featlabel/.style={font=\scriptsize, text=black!60},
]

\node[inputbox, text width=1.6cm] (input) {Input\\$32{\times}32{\times}3$};

\node[convblock, right=0.8cm of input] (conv1) {DW Block 1\\{\scriptsize $3{\times}3$ DW+PW+BN}};
\node[convblock, right=0.5cm of conv1] (conv2) {DW Block 2\\{\scriptsize $3{\times}3$ DW+PW+BN}};
\node[convblock, right=0.5cm of conv2] (conv3) {DW Block 3\\{\scriptsize $3{\times}3$ DW+PW+BN}};

\node[featlabel, below=0.02cm of conv1] {$16{\times}16{\times}C_1$};
\node[featlabel, below=0.02cm of conv2] {$8{\times}8{\times}C_2$};
\node[featlabel, below=0.02cm of conv3] {$h \in \mathbb{R}^d$};

\node[font=\scriptsize, text=blue!60, above=0.08cm of conv2] {$w$-scaled channels};

\draw[thickarr] (input) -- (conv1);
\draw[thickarr] (conv1) -- (conv2) node[midway, above, font=\scriptsize] {pool};
\draw[thickarr] (conv2) -- (conv3) node[midway, above, font=\scriptsize] {pool};

\node[routerbox, below=0.8cm of conv2, text width=5.5cm, minimum height=1.2cm] (router) {
    \textbf{Sparse Router + Hard Capacity}\\[3pt]
    $\tilde{\ell}_i = \underbrace{W_r h}_{\text{base}} + \underbrace{w_k \cos(Kh, e_i)}_{\text{key-sim}}$\\[3pt]
    {\scriptsize Top-$k$ from $\operatorname{softmax}(\tilde{\ell}/\tau)$ \quad $\bullet$ \quad Hard cap: $\lceil c \cdot kB/E \rceil$}
};

\node[utilbox, left=0.8cm of router, text width=1.8cm] (capacity) {
    \textbf{Capacity}\\[1pt]
    {\scriptsize $\text{cap}_i = \lceil c \cdot kB/E \rceil$}\\[1pt]
    {\scriptsize $+\,\mathcal{L}_{\text{lb}},\,\mathcal{L}_{\text{ent}}$}
};

\draw[thickarr] (conv3.east) -- ++(0.3,0) |- (router.east) node[pos=0.25, right, font=\scriptsize] {$h$};

\draw[dasharr, red!70, line width=0.6pt] (capacity) -- (router) node[midway, above, font=\scriptsize, text=red!70] {enforce};

\def\expY{-4.6}
\def\expSpacing{1.35}

\foreach \i/\label in {0/E$_0$, 1/E$_1$, 2/E$_2$} {
    \node[expertbox] at (\i*\expSpacing - 0.5, \expY) (exp\i) {\label\\{\scriptsize MLP $h$}};
}
\node[expertsel] at (3*\expSpacing - 0.5, \expY) (exp3) {E$_3$\\{\scriptsize MLP $h$}};
\node[font=\scriptsize, text=orange!80, anchor=west] at ([xshift=0.25cm, yshift=-0.3cm]exp3.south) {\textbf{selected}};

\foreach \i/\label in {4/E$_4$, 5/E$_5$, 6/E$_6$, 7/E$_7$} {
    \node[expertbox] at (\i*\expSpacing - 0.5, \expY) (exp\i) {\label\\{\scriptsize MLP $h$}};
}

\foreach \i in {0,1,2,4,5,6,7} {
    \draw[arr, black!25] (router.south) -- (exp\i.north);
}
\draw[thickarr, orange!80] (router.south) -- (exp3.north) node[midway, right=2pt, font=\scriptsize, text=orange!70] {$k{=}1$};

\node[outbox, below=0.7cm of exp3, text width=3.0cm] (combine) {
    $o = \sum_i C_i \cdot \text{Exp}_i(h)$\\[2pt]
    {\scriptsize $\rightarrow$ Classifier ($N_c$ classes)}
};

\draw[thickarr, orange!80] ([xshift=0.2cm]exp3.south) -- ([xshift=0.2cm]combine.north);

\coordinate (fb_bottom) at ([xshift=-0.2cm, yshift=-0.5cm]exp3.south);
\coordinate (fb_left) at ([xshift=-1.8cm]capacity.west |- fb_bottom);
\coordinate (fb_top) at (fb_left |- capacity.west);
\draw[dasharr, red!70, line width=0.6pt]
    ([xshift=-0.2cm]exp3.south) -- (fb_bottom) -- (fb_left) -- (fb_top) -- (capacity.west);
\node[font=\scriptsize, text=red!60, right=2pt] at ($(fb_left)!0.65!(fb_top)$) {$\mathcal{L}_{\text{lb}} + \mathcal{L}_{\text{ent}}$};

\node[gabox, right=0.6cm of exp7, text width=2.4cm, minimum height=2.2cm] (ga) {
    \textbf{GA Search}\\[3pt]
    {\scriptsize Pop.\ 6, 3 gen.}\\
    {\scriptsize 14 evals, 12 GPU-hr}\\[4pt]
    {\scriptsize Optimizes:}\\
    {\scriptsize $w$, $c$,}\\
    {\scriptsize $\lambda_{\text{lb}}$, $\lambda_{\text{ent}}$}
};

\coordinate (ga_top) at ([yshift=0.5cm]conv1.north);
\coordinate (ga_vert_top) at (ga.north |- ga_top);
\draw[dasharr, violet!60, line width=0.6pt] (ga.north) -- (ga_vert_top) -- (conv1.north |- ga_top) -- (conv1.north);
\node[font=\scriptsize, text=violet!60, left=2pt] at ($(ga.north)!0.5!(ga_vert_top)$) {$w$};
\draw[dasharr, violet!60, line width=0.6pt] (ga.west) -- (exp7.east);
\draw[dasharr, violet!60, line width=0.6pt, bend right=15] (combine.east) to node[pos=0.5, above, font=\scriptsize, text=violet!60] {fitness} (ga.south west);

\node[font=\small, text=black!60, anchor=north, below=0.3cm of combine] {
    \textcolor{black!70}{$\rightarrow$} Forward pass \quad
    \textcolor{red!70}{- - $\rightarrow$} Aux.\ loss feedback \quad
    \textcolor{violet!60}{- - $\rightarrow$} GA outer loop \quad
    \textcolor{orange!80}{$\boldsymbol{\rightarrow}$} Selected expert ($k{=}1$)
};

\end{tikzpicture}%
}
\caption{Architecture overview. Features from the depthwise backbone are routed by a sparse router combining learned logits and key-similarity. Hard capacity constraints enforce balanced dispatch across $E{=}8$ experts, complemented by load-balance and entropy auxiliary losses. A GA co-optimizes four hyperparameters ($w$, $c$, $\lambda_{\text{lb}}$, $\lambda_{\text{ent}}$) in 14 evaluations.}
\label{fig:architecture}
\end{figure*}
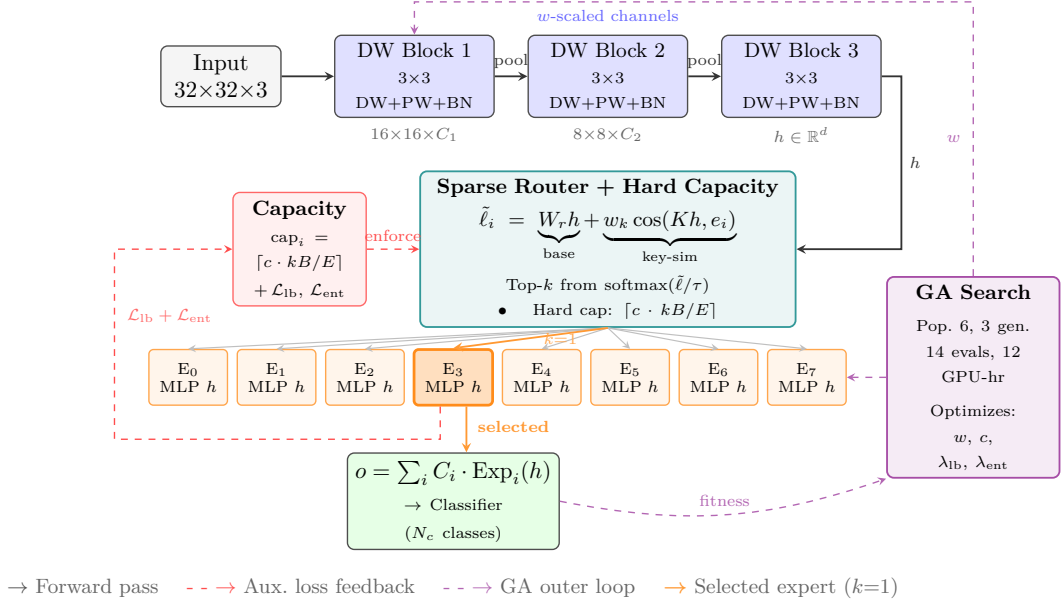

\subsection{Architecture Overview}

Our architecture comprises a shared convolutional backbone feeding either a dense classifier or an MoE layer (Fig.~\ref{fig:architecture}). We study three backbone variants. The \emph{standard backbone} uses two $3{\times}3$ conv blocks (24/48 channels) for CIFAR and three blocks (32/64/128) for Tiny-ImageNet, with BatchNorm on Tiny-ImageNet. The \emph{depthwise backbone} replaces each $3{\times}3$ convolution with a depthwise $3{\times}3$ followed by a pointwise $1{\times}1$, scaled by width factor $w$~\cite{chollet2017xception,howard2017mobilenets}. For ImageNet-1K, we use two pretrained backbones: \emph{ResNet-18}~\cite{he2016deep} (512-dim features) and \emph{MobileNet-V2}~\cite{sandler2018mobilenetv2} (1280-dim features), both after global average pooling.

Experts are MLPs with two hidden layers ($d_{\text{in}} \to h \to h \to d_{\text{out}}$), ReLU activations, dropout 0.3, and hidden widths $h$ set per dataset (304 for CIFAR, 1024 for Tiny-ImageNet, 512 for ImageNet; $h{=}256$ sensitivity in Section~\ref{subsec:imagenet}).

Throughout the paper, we denote the routed share of total inference FLOPs by $\rho = \operatorname{FLOPs}_{\text{routed}} / \operatorname{FLOPs}_{\text{total}}$. For head-only MoE experiments, $\operatorname{FLOPs}_{\text{routed}}$ is the classifier head; for backbone MoE, it is the replaced convolutional block compute.

\subsection{Sparse Routing with Hard Capacity Constraints}

The router produces logits $\ell_i = (W_r h)_i$ from input features $h \in \mathbb{R}^{d_{\text{in}}}$ using a learned routing matrix $W_r \in \mathbb{R}^{E \times d_{\text{in}}}$. A learned key-similarity term provides content-based routing: each expert maintains a key embedding $e_i \in \mathbb{R}^{d_k}$, and a projection $K \in \mathbb{R}^{d_k \times d_{\text{in}}}$ maps input features to the same space ($d_k{=}64$). The combined routing scores are:
\begin{equation}
  \tilde{\ell}_i = \ell_i + w_k \cos(Kh,\, e_i), \quad i = 1,\ldots,E,
  \label{eq:routing}
\end{equation}
where $w_k{=}0.5$ is the key-similarity weight and $\cos(\cdot,\cdot)$ denotes cosine similarity. The top-$k$ experts are selected from $\operatorname{softmax}(\tilde{\ell}/\tau)$, where $\tau$ is the temperature. The routing schedule anneals temperature from $\tau_{\max}$ to $\tau_{\min}$ using either a linear schedule $\tau(t) = \tau_{\max} - (\tau_{\max} - \tau_{\min})\,t/T$ or a sigmoid schedule:
\begin{equation}
  \tau(t) = \tau_{\min} + (\tau_{\max} - \tau_{\min}) \cdot \sigma(-\kappa(t/T - 0.5)),
  \label{eq:sigmoid-schedule}
\end{equation}
where $\sigma$ is the logistic function and $\kappa{=}7.0$ the sharpness. Per-dataset schedule choices are specified in Section~\ref{subsec:setup}.

While load-balancing ($\lambda_{\text{lb}}$) and entropy ($\lambda_{\text{ent}}$) regularizers encourage uniform expert usage, they do not guarantee balanced per-batch dispatch. Following Switch Transformer~\cite{fedus2022switch}, we therefore enforce hard capacity limits:
\begin{equation}
  \text{cap}_i = \left\lceil c \cdot \frac{k \cdot B}{E} \right\rceil, \quad \forall i \in \{1,\ldots,E\},
  \label{eq:capacity}
\end{equation}
where $c$ is the capacity factor, $B$ the batch size, $k$ the top experts selected, and $E$ the total experts. Samples are assigned to their highest-scoring expert with available capacity; if all $k$ candidates are full, the sample is force-assigned to its top-1 expert (with $c{>}1$, this fallback is rare). The dispatch mask $M \in \{0,1\}^{B \times E}$ records assignments. Let $p_{b,:} = \operatorname{softmax}(\tilde{\ell}_b/\tau)$; the combine weights retain only the probabilities of assigned experts, renormalized: $C_{b,i} = M_{b,i}\, p_{b,i} / \sum_j M_{b,j}\, p_{b,j}$, so that $o_b = \sum_i C_{b,i}\, \text{Expert}_i(h_b)$.

\textbf{Development note.} During development, we also explored an additive utility bias $\lambda_u u_i$ added to the routing scores of Eq.~\ref{eq:routing}, where $u_i$ is an EMA of expert gradient magnitudes intended to reward actively learning experts. Code inspection revealed that this bias was operationally negligible in our CIFAR experiments: the additive term was too small relative to learned router margins to change any top-$k$ selections. We retain the utility machinery in the released code for reproducibility but do not claim it as a contribution.

\subsection{Training Objective}

The total loss combines cross-entropy with two regularizers:
\begin{equation}
  \mathcal{L} = \mathcal{L}_{\text{CE}} + \lambda_{\text{lb}} \mathcal{L}_{\text{lb}} + \lambda_{\text{ent}} \mathcal{L}_{\text{ent}},
\end{equation}
where $\mathcal{L}_{\text{lb}}$ is the Switch Transformer load-balance loss~\cite{fedus2022switch}:
\begin{equation}
  \mathcal{L}_{\text{lb}} = E \cdot \sum_{i=1}^{E} f_i \cdot p_i,
\end{equation}
with $f_i$ the fraction of samples dispatched to expert $i$ and $p_i$ the mean routing probability for expert $i$. The entropy regularizer $\mathcal{L}_{\text{ent}} = -H(\bar{p})$, where $\bar{p}$ is the batch-averaged routing probability vector, penalizes concentrated routing distributions.

Algorithm~\ref{alg:training} summarizes one training iteration.

\begin{algorithm}[t]
  \caption{Sparse MoE training with hard capacity (one iteration)}
  \label{alg:training}
  \begin{algorithmic}[1]
    \STATE $(\tau, k) \gets \textsc{Schedule}(\text{epoch})$
    \STATE Sample mini-batch $(x, y)$
    \STATE $h \gets \textsc{Backbone}(x)$
    \STATE $\tilde{\ell}_i \gets (W_r h)_i + w_k \cos(Kh,\, e_i)$ for $i=1,\ldots,E$ \COMMENT{Routing scores}
    \STATE $(M, C) \gets \textsc{CapacityAwareTopK}(\operatorname{softmax}(\tilde{\ell}/\tau), k, c)$ \COMMENT{Mask, combine weights}
    \STATE $o \gets \textsc{DispatchAndCombine}(h, \{\text{Exp}_i\}, M, C)$
    \STATE $\mathcal{L} \gets \mathcal{L}_\text{CE}(o, y) + \lambda_{\text{lb}} \mathcal{L}_{\text{lb}} + \lambda_{\text{ent}} \mathcal{L}_{\text{ent}}$
    \STATE Backpropagate and update all parameters
  \end{algorithmic}
\end{algorithm}

\subsection{Evolutionary Search and Baselines}
\label{sec:evo-search}

Four hyperparameters govern the accuracy--efficiency trade-off: width scale $w$, capacity factor $c$, load-balance weight $\lambda_{\text{lb}}$, and entropy weight $\lambda_{\text{ent}}$ (search ranges and optimized values in supplementary Section~D). We designate CIFAR-10 as the \emph{development dataset} for hyperparameter search; consequently, CIFAR-10 results may be optimistic and our strongest generalization evidence rests on the three transfer datasets (CIFAR-100, Tiny-ImageNet, ImageNet-1K). The three routing hyperparameters ($c$, $\lambda_{\text{lb}}$, $\lambda_{\text{ent}}$) are held fixed across datasets. Dataset-specific training parameters (warmup epochs, $\tau_{\min}$, expert hidden dimension $h$) are adapted per dataset (Section~\ref{subsec:setup}). We use population size 6 with top-2 elites; each candidate trains for 50 epochs with fitness
$F = \text{Acc}_{\text{test}} - \lambda_{\text{red}} \max(0, r^* - r) - \lambda_{\text{gap}} \max(0, |g| - g^*)$,
where $r$ is the observed FLOPs reduction, $g$ the MoE--dense gap, $r^*{=}0.2$, $g^*{=}0.02$, $\lambda_{\text{red}}{=}5$, $\lambda_{\text{gap}}{=}2$. The search converged in fourteen evaluations (${\sim}$12 GPU-hours); the narrow ranges constrain the optimizable volume, and the three routing hyperparameters transfer to all three evaluation datasets without retuning.

For cross-method comparison, we reimplement Soft MoE~\cite{puigcerver2024soft} and Expert Choice~\cite{zhou2022expertchoice}, adapting each to our per-sample CNN setting. All MoE variants share the same expert architecture (8 experts, 2-hidden-layer MLP) and base training configuration to ensure fair comparison.

\subsection{Per-Sample Soft Gating}
\label{subsec:persample-soft}

Soft MoE~\cite{puigcerver2024soft} dispatches via softmax over the batch dimension, which averages unrelated images in per-sample CNN classification. To isolate this as the collapse cause, we implement \emph{per-sample soft gating}: each expert has a slot embedding $s_i \in \mathbb{R}^{d}$, and gating weights are:
\begin{equation}
  w_i = \frac{\exp(h^\top s_i / \tau)}{\sum_{j=1}^{E} \exp(h^\top s_j / \tau)}, \quad i = 1,\ldots,E,
  \label{eq:persample-gate}
\end{equation}
where the softmax is over experts (dim=$E$), not over the batch. The combined output is:
\begin{equation}
  o = \sum_{i=1}^{E} w_i \cdot \text{Expert}_i(h).
  \label{eq:persample-combine}
\end{equation}
This preserves per-sample discriminative information while retaining Soft MoE's differentiable, token-drop-free dispatch. No auxiliary load-balancing loss is needed since all experts process every sample.

\section{Experiments}
\label{sec:experiments}

We validate the sparse MoE architecture on four vision benchmarks of increasing complexity, first using CIFAR-10 as a development set for architecture search, then transferring the optimized routing configuration to three progressively harder evaluation datasets.

\subsection{Experimental Setup}
\label{subsec:setup}

\textbf{Datasets.} We evaluate on four datasets: CIFAR-10 and CIFAR-100~\cite{krizhevsky2009learning} ($32{\times}32$, 50k/10k train/test), Tiny-ImageNet (200 classes, $64{\times}64$, 100k train), and ImageNet-1K~\cite{deng2009imagenet} (1000 classes, $224{\times}224$, 1.28M/50k train/val).

\textbf{Training.} CIFAR trains for 50 epochs, Tiny-ImageNet for 80, ImageNet for 30, using Adam~\cite{kingma2015adam} (lr=$10^{-3}$, CIFAR/Tiny-IN) or SGD (lr=$10^{-2}$, momentum 0.9, cosine schedule, ImageNet). Pretrained ImageNet backbones are fine-tuned at $0.1\times$ head lr with AMP BF16 and batch 256. We use standard augmentations (flips, random crops) rather than modern recipes (AdamW, CutMix) to isolate routing effects; dense and MoE models share identical training (see Limitations). FLOPs are computed analytically.

\textbf{MoE Schedule.} Temperature anneals from $\tau{=}1.0$ to $\tau_{\min}$ (sigmoid $\kappa{=}7.0$ for CIFAR $w{=}2.0$; linear otherwise; $\tau_{\min}{=}0.13$ CIFAR, $0.3$ Tiny-IN/ImageNet). CIFAR uses 5-epoch warmup ($k{=}8{\to}1$); other datasets use $k{=}1$ throughout.

\textbf{Baselines.} We compare four routing methods: (i)~Dense (FC head), (ii)~Sparse MoE $k{=}1$ with hard capacity constraints (Ours), (iii)~Soft MoE~\cite{puigcerver2024soft}, and (iv)~Expert Choice~\cite{zhou2022expertchoice}. All MoE variants share expert architecture (8 experts, 2-hidden-layer MLP).

\textbf{Statistical Validation.} Headline results report mean$\pm$s.d.\ over 10 seeds (CIFAR) or 5 seeds (Tiny-ImageNet/ImageNet) with paired $t$-tests, Cohen's $d$, and 95\% CIs. CIFAR reports final-epoch test accuracy; Tiny-ImageNet and ImageNet report peak validation accuracy (more reliable under higher training variance).

\textbf{Search Protocol Disclosure.} CIFAR-10 serves as the hyperparameter development set: evolutionary search fitness is evaluated on the CIFAR-10 \emph{test split}, so CIFAR-10 results may be optimistic and should be treated as development evidence rather than generalization evidence. Our strongest generalization claims rest on the three transfer datasets (CIFAR-100, Tiny-ImageNet, ImageNet-1K), where no test-set information informed the architectural or routing hyperparameter search.

\textbf{Comparison Structure.} Cross-method tables use a two-tier design: the \emph{standard backbone} section isolates routing effects on identical architectures; the \emph{depthwise} section reports the full co-optimization recipe. We note that Soft MoE and Expert Choice are evaluated only on standard backbones; their behavior under depthwise co-optimization remains unexplored, so the cross-method comparison speaks to routing-method quality on matched architectures, not to the full recipe. The controlled $\rho$-sweep in Table~\ref{tab:c1-rho-sweep} shows that neither backbone family alone nor sparse routing at low $\rho$ explains the gains. Single-seed rows are exploratory; all claims rest on multi-seed results. Supplementary Section~E contextualizes absolute accuracy against published baselines.

\subsection{CIFAR-10: Architecture Search}
\label{subsec:cifar10}

We employ a lightweight CNN backbone to isolate sparse MoE routing effects, focusing on \emph{relative improvement} over matched dense baselines. Development sweeps over backbone width, expert hidden dimension, temperature schedule, and routing variants (full table in supplementary Section~H) yield two headline depthwise-MoE configurations: an efficiency-oriented variant ($w{=}0.72$) achieving $+1.28 \pm 1.26\%$ over dense at a $22.7\%$ FLOPs reduction ($p{=}.011$, $d{=}1.01$), and a higher-accuracy variant ($w{=}2.0$, three conv blocks with BatchNorm) reaching $88.41\%$ test accuracy with $+0.54 \pm 0.16\%$ over dense ($p{<}10^{-5}$, $d{=}3.31$, all 10 seeds positive). The two sit at different accuracy--efficiency operating points on the same recipe, both with significant positive gaps.

\textbf{Cross-Method Comparison.} On CIFAR-10, both Soft MoE and Expert Choice underperform the dense baseline on the standard backbone; only our depthwise co-optimized configuration achieves a positive gap. Cross-method numbers across datasets are consolidated in Table~\ref{tab:cross-dataset-methods} (Section~\ref{sec:analysis}).

\subsection{Controlled \texorpdfstring{$\rho$}{rho}-sweep on CIFAR-10}
\label{subsec:rho-sweep}

To test whether $\rho$ (the routed share of total FLOPs), rather than backbone family, is the active variable behind the compute-leverage observation, we sweep expert hidden size $h \in \{128, 512, 2048\}$ across both standard and depthwise backbones (six configurations $\times$ five seeds $=$ 30 runs). The dense head is $\text{feature\_dim} \to 1024 \to h \to 10$ and each expert is $\text{feature\_dim} \to h \to h \to 10$, sharing $h$ to remove the fixed-256 mismatch in earlier experiments. Two predictions follow: a high-$\rho$ \emph{standard} backbone should turn positive, and a low-$\rho$ \emph{depthwise} backbone should turn negative.

\begin{table}[t]
  \centering
  \caption{CIFAR-10 $\rho$-sweep (final-epoch test accuracy, 5-seed mean$\pm$s.d.). Both predicted sign reversals (highlighted) hold. Per-seed numbers in supplementary Section~G.}
  \label{tab:c1-rho-sweep}
  \footnotesize\setlength{\tabcolsep}{3pt}
  \begin{tabular}{@{}lcccccc@{}}
    \toprule
    Config & $\rho$ & Dense & MoE & Gap & sd & $t$ \\
    \midrule
    Std.\ $h{=}128$   & 5.5\%  & 84.34 & 82.16 & $-2.18$ & 0.47 & $-10.34$ \\
    DW $h{=}128$      & 18.0\% & 76.75 & 75.48 & \textbf{$-1.27$} & 2.27 & $-1.25$ \\
    Std.\ $h{=}512$   & 19.7\% & 83.51 & 83.24 & $-0.27$ & 0.62 & $-0.99$ \\
    DW $h{=}512$      & 48.8\% & 76.25 & 78.16 & $+1.90$ & 2.03 & $+2.10$ \\
    Std.\ $h{=}2048$  & 54.7\% & 83.04 & 84.11 & \textbf{$+1.07$} & 1.10 & $+2.17$ \\
    DW $h{=}2048$     & 83.3\% & 75.90 & 80.69 & $+4.79$ & 1.79 & $+6.00$ \\
    \bottomrule
  \end{tabular}
\end{table}

Table~\ref{tab:c1-rho-sweep} reports the result. Gap is monotonic in $\rho$ across all six configurations, and both predicted reversals are observed: \texttt{Std.\ $h{=}2048$} flips positive and \texttt{DW $h{=}128$} flips negative. This refutes both ``depthwise is required'' and ``depthwise alone is sufficient,'' and is the cleanest evidence we have that $\rho$, not backbone family, is the active variable.

We note one caveat: the sweep varies head capacity and $\rho$ jointly, since the expert hidden size determines both the routed FLOP fraction and the per-expert capacity. The \emph{direction} of effect cleanly tracks $\rho$; the \emph{magnitude} mixes capacity. A strict $\rho$-isolation experiment with FLOP-matched dense-head controls is left to future work.

\subsection{Multi-Dataset Scaling}
\label{subsec:scaling}

Having established routing effectiveness on the development benchmark, we transfer the four GA-optimized routing hyperparameters to three progressively more complex datasets to test generalization.

\textbf{CIFAR-100.} The MoE--dense gap widens on the more complex CIFAR-100 benchmark (Table~\ref{tab:cifar100-results}). The GA-optimized routing hyperparameters transfer without retuning, achieving a positive gap even in zero-shot mode; all seeds are positive ($p{=}0.0013$). Notably, the wider DW+MoE model exceeds the standard dense baseline in absolute accuracy, demonstrating that the depthwise efficiency penalty need not compromise performance.

\begin{table}[t]
  \centering
  \caption{CIFAR-100 results (\%, final-epoch test accuracy). %
  Std backbone is single-seed reference. %
  \textsuperscript{$\dagger$}Zero-shot: CIFAR-10 optimized config without retuning. %
  \textsuperscript{$\ddagger$}10-seed mean$\pm$s.d.}
  \label{tab:cifar100-results}
  \footnotesize\setlength{\tabcolsep}{2.5pt}
  \begin{tabular}{@{}lcccc@{}}
    \toprule
    Configuration & Dense & MoE & Gap & $\Delta$FLOPs \\
    \midrule
    Std backbone (24/48) & 57.93 & 53.90 & $-4.03$ & $-$8.8\% \\
    \midrule
    Opt.\ DW Slim (ZS)\textsuperscript{$\dagger$} & 43.72 & 44.02 & $+0.30$ & $-$22.7\% \\
    Opt.\ DW Wide (single) & 44.11 & 46.98 & $+2.87$ & $-$18.7\% \\
    \midrule
    \multicolumn{5}{@{}l}{\emph{10-seed (Opt.\ DW Slim)\textsuperscript{$\ddagger$}:}} \\
    Mean$\pm$s.d. & 41.38$\pm$2.10 & 44.37$\pm$1.61 & $\mathbf{+2.99\pm2.07}$ & $-$22.7\% \\
    \multicolumn{5}{@{}l}{\quad $p{=}.0013$, $d{=}1.45$, 95\% CI $[+1.51, +4.47]$} \\
    \midrule
    \multicolumn{5}{@{}l}{\emph{10-seed (3-blk+BN DW $w{=}2.0$)\textsuperscript{$\ddagger$}:}} \\
    Mean$\pm$s.d. & 59.85$\pm$0.56 & 61.29$\pm$0.88 & $+1.44\pm1.06$ & $-$4.3\% \\
    \multicolumn{5}{@{}l}{\quad $p{=}.0019$, $d{=}1.37$, 95\% CI $[+0.69, +2.20]$} \\
    \bottomrule
  \end{tabular}
\end{table}

\textbf{CIFAR-100 Cross-Method Comparison.} On CIFAR-100, Soft MoE's token-mixing dispatch collapses catastrophically because each ``token'' is an independent image rather than a semantically related ViT~\cite{dosovitskiy2021image} patch. Per-sample soft gating (Section~\ref{subsec:persample-soft}) eliminates this collapse, recovering above the dense baseline; Expert Choice avoids the dispatch mismatch but still trails dense. Numbers in Table~\ref{tab:cross-dataset-methods}.

\textbf{Tiny-ImageNet.} Extending to 200 classes at $64{\times}64$ (Table~\ref{tab:tinyimagenet-results}), the MoE--dense gap increases further. The best configuration ($w{=}1.2$) achieves the largest gap across all datasets ($p{<}10^{-4}$, $d{=}9.52$), reflecting remarkably low seed-to-seed variance. Critically, DW+MoE \emph{exceeds} the standard dense baseline in absolute accuracy at roughly 61\% of standard-backbone FLOPs. Positive $\Delta$FLOPs values reflect that large experts exceed the dense head cost (Section~\ref{subsec:flops-decomp}).

\begin{table}[t]
  \centering
  \caption{Tiny-ImageNet results (\%). Peak = best validation accuracy during training. %
  \textsuperscript{$\ddagger$}5-seed mean$\pm$s.d. DW $w$ = depthwise width scale.}
  \label{tab:tinyimagenet-results}
  \footnotesize\setlength{\tabcolsep}{2.5pt}
  \begin{tabular}{@{}lcccc@{}}
    \toprule
    Config & Dense & MoE Peak & Gap & $\Delta$FLOPs \\
    \midrule
    Std ($h{=}1024$, 1-seed) & 43.73 & 46.33 & $+2.60$ & $+$5.5\% \\
    \midrule
    \multicolumn{5}{@{}l}{\emph{5-seed (DW $w{=}0.72$, $h{=}1024$)\textsuperscript{$\ddagger$}:}} \\
    Mean$\pm$s.d. & 39.23$\pm$.46 & 42.61$\pm$.30 & $+3.38\pm.48$ & $+$18.2\% \\
    \multicolumn{5}{@{}l}{\quad $p{<}10^{-4}$, $d{=}7.00$, CI $[+2.78, +3.97]$} \\
    \midrule
    \multicolumn{5}{@{}l}{\emph{5-seed (DW $w{=}1.2$, $h{=}1024$)\textsuperscript{$\ddagger$}:}} \\
    Mean$\pm$s.d. & 42.36$\pm$.20 & \textbf{46.35$\pm$.54} & $\mathbf{+3.99\pm.42}$ & $+$12.0\% \\
    \multicolumn{5}{@{}l}{\quad $p{<}10^{-4}$, $d{=}9.52$, CI $[+3.47, +4.51]$} \\
    \midrule
    \multicolumn{5}{@{}l}{\emph{5-seed (DW $w{=}2.0$, $h{=}1024$)\textsuperscript{$\ddagger$}:}} \\
    Mean$\pm$s.d. & 44.05$\pm$.40 & 45.86$\pm$1.77 & $+1.81\pm1.89$ & $+$7.8\% \\
    \multicolumn{5}{@{}l}{\quad $p{=}.10$, $d{=}0.95$, CI $[-0.54, +4.15]$} \\
    \bottomrule
  \end{tabular}
\end{table}

\textbf{Tiny-ImageNet Cross-Method Comparison.} On the standard backbone, Soft MoE's token-mixing dispatch collapses to near-random performance, while per-sample gating recovers to near-dense levels. Expert Choice and our own sparse MoE both show small negative gaps on the standard backbone---underscoring that the positive multi-seed results in Table~\ref{tab:tinyimagenet-results} arise from the joint DW co-optimization recipe, not from routing alone. Numbers in Table~\ref{tab:cross-dataset-methods}.

\begin{figure}[t]
  \centering
  \includegraphics[width=0.57\linewidth]{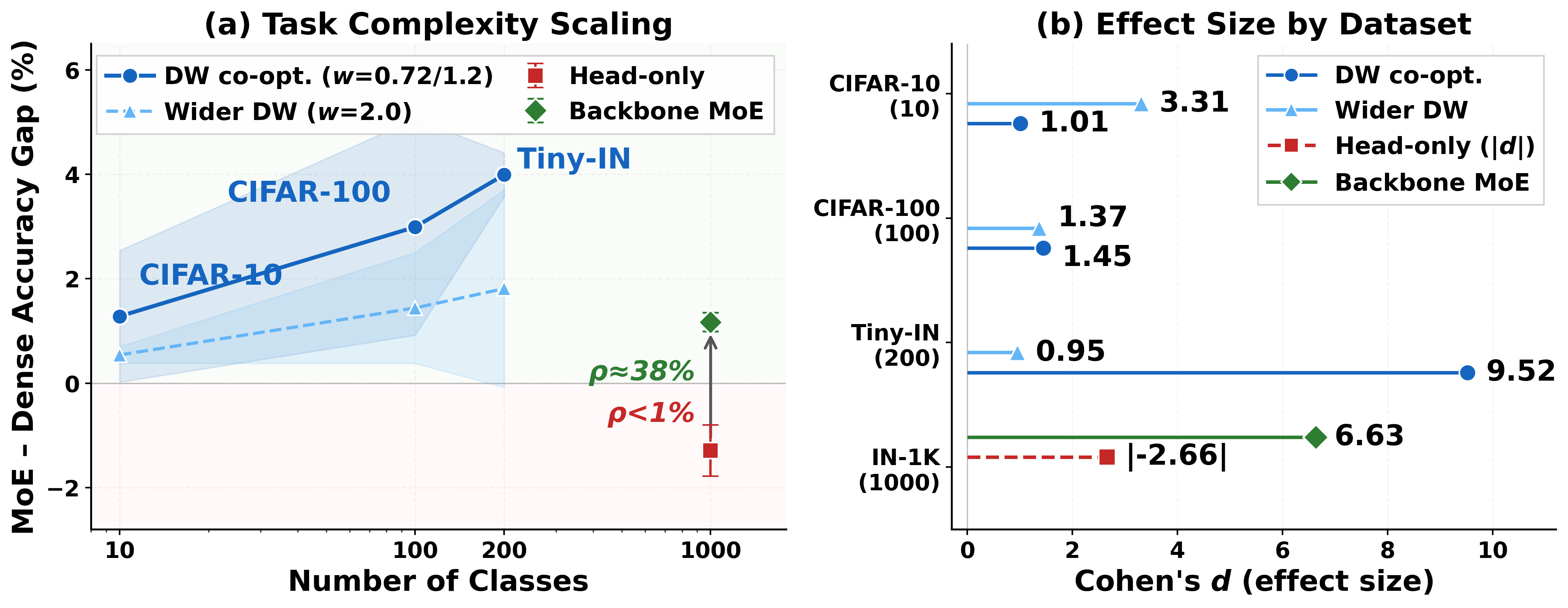}
  \caption{Task complexity scaling. Left: MoE--dense gap vs.\ number of classes. Right: Cohen's $d$ effect size. Green/red shading marks positive/negative regimes. The two ImageNet-1K points are the head-only and backbone-MoE configurations (Section~\ref{subsec:backbone-moe}).}
  \label{fig:scaling-trend}
\end{figure}

\textbf{Task Complexity Scaling.} Fig.~\ref{fig:scaling-trend} summarizes the multi-seed gaps reported in Tables~\ref{tab:cifar100-results}, \ref{tab:tinyimagenet-results}, and \ref{tab:imagenet-results}. Under depthwise co-optimization, the gap trends upward with class count across the three smaller-scale datasets, with the caveat that these comparisons span different backbone widths and evaluation protocols (final-epoch test accuracy for CIFAR vs.\ peak validation for Tiny-ImageNet/ImageNet). The controlled $w{=}2.0$ series shows the same monotonic pattern on matched architectures, though Tiny-ImageNet does not reach significance ($p{=}0.10$). ImageNet-1K appears as two distinct configurations: head-only MoE yields negative gaps on both backbones, while backbone MoE recovers a positive gain (Section~\ref{subsec:backbone-moe}); the joint roles of $\rho$ and routing regime in this recovery are disentangled by the $k$-ablation in that section.

A utility-bias ablation in the supplementary material confirms that the additive utility term explored during development does not explain the observed gains.

\subsection{ImageNet-1K Validation}
\label{subsec:imagenet}

To test whether MoE advantages extend to large-scale classification, we evaluate on ImageNet-1K using two pretrained backbones: ResNet-18 ($\rho{=}0.06\%$) and MobileNet-V2 ($\rho{=}0.85\%$). Both have $\rho < 1\%$, so these experiments test the compute-leverage hypothesis.

\begin{table}[t]
  \centering
  \caption{ImageNet-1K results (peak validation accuracy, 5-seed mean$\pm$s.d.). %
  \textsuperscript{$\star$}Per-sample soft gating. %
  \textsuperscript{$\dagger$}Token-mixing Soft MoE (3-seed).}
  \label{tab:imagenet-results}
  \footnotesize\setlength{\tabcolsep}{2.5pt}
  \begin{tabular}{@{}llcccc@{}}
    \toprule
    Backbone & Method & Top-1 & Gap & $h$ & $\rho$ (rho) \\
    \midrule
    \multirow{5}{*}{ResNet-18} & Dense (FC) & $70.29\pm0.04$ & --- & --- & \multirow{5}{*}{0.06\%} \\
    & MoE $k{=}1$ & $68.99\pm0.51$ & $-1.29$ & 512 \\
    & Soft (token)\textsuperscript{$\dagger$} & $41.03\pm0.24$ & $-29.25$ & 512 \\
    & \textbf{Soft (per-samp.)}\textsuperscript{$\star$} & $\mathbf{69.67\pm0.08}$ & $\mathbf{-0.66}$ & 512 \\
    & Expert Choice & $41.25\pm1.72$ & $-29.03$ & 512 \\
    \midrule
    \multirow{3}{*}{MobileNet-V2} & Dense (FC) & $72.26\pm0.06$ & --- & --- & \multirow{3}{*}{0.85\%} \\
    & MoE $k{=}1$ ($h{=}512$) & $70.35\pm0.45$ & $-1.90$ & 512 \\
    & MoE $k{=}1$ ($h{=}256$) & $68.07\pm0.55$ & $-4.19$ & 256 \\
    \midrule
    \multirow{2}{*}{\shortstack[l]{ResNet-18\\(backbone)}} & Dense (layer3/4) & $70.04\pm0.14$ & --- & --- & \multirow{2}{*}{$\approx$38\%} \\
    & \textbf{MoE $k{=}2$}\textsuperscript{$\ddagger$} & $\mathbf{71.21\pm0.17}$ & $\mathbf{+1.17}$ & --- \\
    \bottomrule
    \multicolumn{6}{@{}l}{\scriptsize\textsuperscript{$\ddagger$}MoEConv2d in layer3/4 ($E{=}8$); details in Section~\ref{subsec:backbone-moe}.}
  \end{tabular}
\end{table}

\textbf{ResNet-18 ($\rho{=}0.06\%$).} Our sparse MoE trails the dense baseline significantly ($p{=}0.004$), consistent with negative margins on standard backbones at smaller scales.

\textbf{MobileNet-V2 ($\rho{=}0.85\%$).} MobileNet-V2 uses depthwise separable convolutions throughout, raising $\rho$ by $14\times$ over ResNet-18. If backbone type were the determining factor, this depthwise backbone should yield positive gaps. Instead, the gap remains significantly negative at both expert sizes (both $p{<}0.001$; Table~\ref{tab:imagenet-results}). This result suggests that $\rho$ is an important factor: depthwise convolutions alone appear insufficient when the head constitutes $<$1\% of total FLOPs. The worsening gap at smaller expert size further confirms that undersized experts amplify the deficit (cf.\ Section~\ref{subsec:hard-capacity}).

\textbf{Per-Sample Soft MoE.} Replacing token-mixing dispatch with per-sample gating (Section~\ref{subsec:persample-soft}) eliminates the catastrophic collapse, recovering the majority of the deficit. The per-sample variant approaches but does not surpass the dense baseline on any dataset (Table~\ref{tab:imagenet-results}), suggesting that weighted expert ensembling without hard routing cannot fully exploit the pretrained feature space.

\textbf{Pretrained Feature Mismatch.}
Token-mixing Soft MoE and Expert Choice both exhibit severe performance degradation despite a strong dense baseline. Epoch-1 accuracy provides insight: the dense model immediately achieves high accuracy because the pretrained features are already linearly separable, whereas the dispatch mechanisms disrupt this separability by mixing unrelated samples before expert processing. Section~\ref{sec:analysis} analyzes the mechanisms underlying these results.

\subsection{MoE-in-Backbone Validation}
\label{subsec:backbone-moe}

The head-only experiments above confirm that routing gains vanish when $\rho < 1\%$. We now test the converse by placing MoE directly within backbone convolutions where the majority of computation resides.

\textbf{Architecture.} We replace ResNet-18's layer3 and layer4 $3{\times}3$ convolutions with MoEConv2d layers ($E{=}8$ expert filter banks, $k{=}2$ routing). The router applies global average pooling followed by a linear layer to produce per-sample expert logits. Early layers (conv1, layer1, layer2) remain frozen pretrained; MoE experts are initialized from pretrained layer3/4 weights with $10\%$ noise perturbation to break symmetry. This places approximately $38\%$ of total inference FLOPs under routing control, well above the ${\sim}7\%$ threshold identified in Section~\ref{subsec:flops-decomp}.

\textbf{Results.} The backbone MoE variant achieves a statistically significant positive gap over the matched dense baseline ($p{=}1.2{\times}10^{-4}$, all five seeds positive; full per-seed accuracies in supplementary Section~G) and is reported alongside the $k{=}1$ ablation in Table~\ref{tab:c3-k1-ablation}. Routing exhibits balanced yet non-uniform expert utilization across all eight MoE layers---all experts remain active with differentiated usage patterns---in contrast to the $k{=}1$ head-only configuration, which converged to exactly uniform routing with no specialization.

The contrast with head-only MoE on the \emph{same backbone} is suggestive: a significant negative gap at $\rho{=}0.06\%$ (Table~\ref{tab:imagenet-results}) versus a significant positive gap once MoE moves into the backbone. This is consistent with $\rho$ being an important factor in routing effectiveness. However, we note that the backbone MoE also differs in routing regime ($k{=}2$ vs.\ $k{=}1$), trainable scope, and initialization strategy.

\textbf{Routing-regime ablation ($k{=}1$ vs.\ $k{=}2$).} We isolate the contribution of multi-expert routing by repeating the backbone-MoE experiment with $k{=}1$ instead of $k{=}2$, holding the architecture, routed fraction, pretrained initialization, and all other hyperparameters fixed. As Table~\ref{tab:c3-k1-ablation} shows, $k{=}1$ flips the gap negative across all five seeds---a 3.25\% reversal at identical $\rho$ that demonstrates multi-expert routing ($k \geq 2$) is necessary in addition to a substantial $\rho$.

\begin{table}[t]
  \centering
  \caption{Routing-regime ablation on backbone MoE (ImageNet-1K, 5-seed mean$\pm$s.d.; same backbone, same routed fraction, same init). $k$ alone reverses the gap by $3.25\%$.}
  \label{tab:c3-k1-ablation}
  \small
  \begin{tabular}{@{}lcccc@{}}
    \toprule
    Routing & Dense & MoE & Gap & $t$ \\
    \midrule
    $k{=}2$ (paper) & $70.04\pm0.14$ & $71.21\pm0.17$ & $\mathbf{+1.17\pm0.18}$ & $+14.82$ \\
    $k{=}1$ (ablation) & $70.03\pm0.05$ & $67.95\pm0.12$ & $\mathbf{-2.08\pm0.15}$ & $-30.76$ \\
    \bottomrule
  \end{tabular}
\end{table}

\section{Mechanistic Analysis}
\label{sec:analysis}

The experiments in Section~\ref{sec:experiments} establish that sparse MoE routing with depthwise backbones achieves consistent gains when the routed FLOP fraction is sufficiently high, and that depthwise backbones help primarily by raising this fraction (Tables~\ref{tab:c1-rho-sweep} and~\ref{tab:flops-decomp}). We now analyze the mechanisms underlying these gains and identify practical design guidelines.

\subsection{Routing Stability}
\label{subsec:hard-capacity}

Temperature scheduling significantly affects routing stability. On the CIFAR-10 development set, aggressive linear annealing to low $\tau_{\min}$ causes late-stage routing collapse where accuracy drops well below the training peak. Sigmoid schedules avoid this collapse; alternatively, raising $\tau_{\min}$ prevents collapse with linear schedules. See supplementary Section~C for the full ablation table.

\begin{figure}[t]
  \centering
  \includegraphics[width=0.665\linewidth]{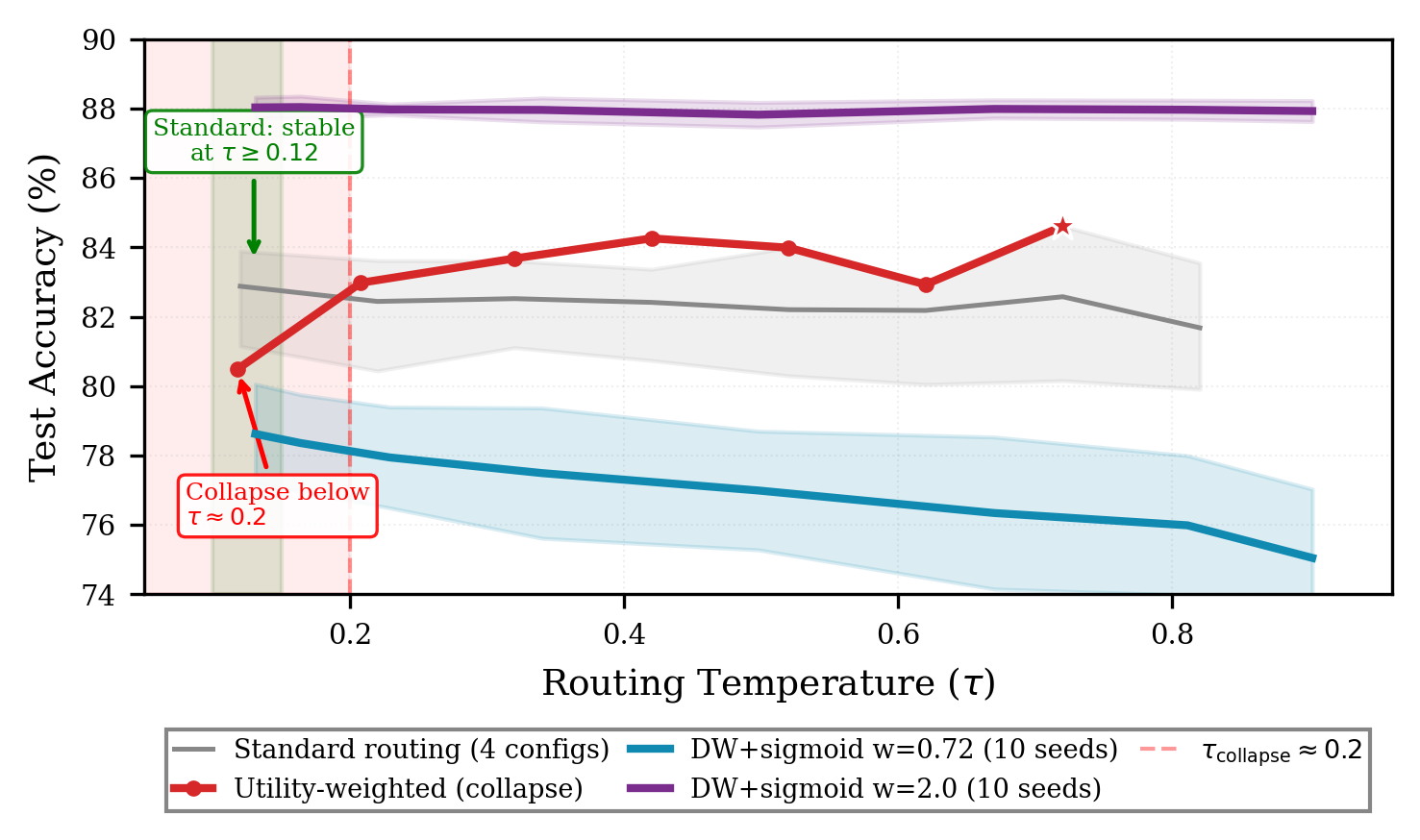}
  \caption{Temperature--accuracy trajectories ($k{=}1$, CIFAR-10). Standard routing (gray) is stable to low $\tau$; routing with additive utility bias (red) collapses under aggressive annealing. Sigmoid schedules (blue/purple, 10 seeds) avoid collapse. The utility bias was operationally negligible (supplementary Section~I); the collapse reflects temperature sensitivity, not the bias itself.}
  \label{fig:temperature-collapse}
\end{figure}

Fig.~\ref{fig:temperature-collapse} illustrates this sensitivity. Prior work has addressed routing collapse through auxiliary losses~\cite{fedus2022switch,zoph2022stmoe}; our experiments show that temperature scheduling is a complementary and important stability lever.

In development runs, soft auxiliary losses alone did not reliably prevent concentrated dispatch, but we did not run a controlled hard-vs-soft capacity ablation. We therefore treat hard capacity as a design constraint rather than as an isolated contribution: Eq.~\ref{eq:capacity} enforces a per-batch ceiling that keeps all experts active during training.

Expert hidden dimension further affects the capacity--diversity trade-off: moderate hidden sizes maintain full diversity, while smaller experts collapse at $k{=}1$ on harder tasks, suggesting the minimum viable expert dimension scales with output complexity.

\subsection{Expert Specialization}

\begin{figure}[t]
  \centering
  \includegraphics[width=0.57\linewidth]{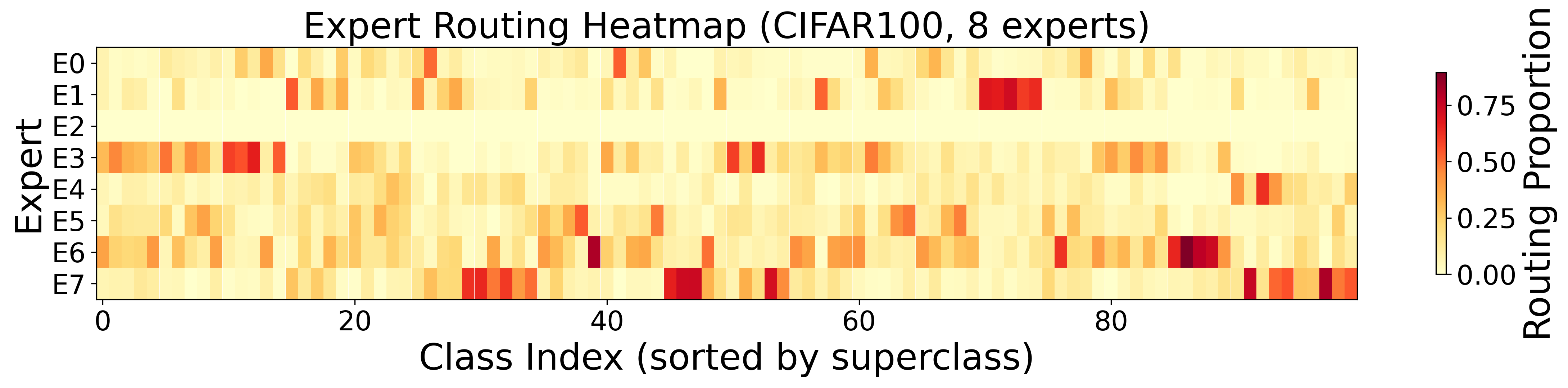}
  \caption{Expert routing heatmap, CIFAR-100 test set (3-block+BN DW $w{=}2.0$, seed 123). Columns sorted by 20 superclasses; seven experts show superclass-aligned specialization.}
  \label{fig:routing-heatmap}
\end{figure}

The CIFAR-100 routing heatmap (Fig.~\ref{fig:routing-heatmap}) shows broad superclass-aligned patterns across multiple active experts; the CIFAR-10 heatmap (supplementary Section~A) produces sharper class specialization, consistent with the lower output complexity. On the narrower backbone ($w{=}0.72$), test-time routing concentrates on 1--3 experts despite training-time usage showing 8/8 active, yet accuracy gaps \emph{increase} with this concentration. A plausible explanation is a \emph{training-time diversity regularization} effect, in which hard capacity constraints force all experts to develop competence during training, building latent capacity that benefits the dominant test-time experts. Training dynamics across all three smaller-scale datasets and a $t$-SNE visualization of expert assignments versus class labels appear in supplementary Sections~B and~A respectively; both are visualizations of the same routing structure and we treat their alignment as qualitative rather than as a quantitative claim of semantic specialization.

\subsection{Efficiency Analysis}
\label{subsec:flops-decomp}

Table~\ref{tab:flops-decomp} decomposes FLOPs into backbone and head for efficiency-oriented expert sizes. On CIFAR, the head accounts for nearly half of total FLOPs, enabling substantial MoE reduction. On Tiny-ImageNet, convolutions dominate; headline configurations use larger experts for accuracy rather than efficiency. On ImageNet, the head is negligible. This decomposition explains the multi-dataset gap pattern: depthwise backbones shift $\rho$ from negligible to near-parity, which is where positive accuracy gaps appear.

\begin{table}[t]
  \centering
  \caption{FLOPs decomposition ($\rho$ = routed share of total FLOPs). %
  Efficiency-oriented $h$: 304 CIFAR, 440 Tiny-IN.}
  \label{tab:flops-decomp}
  \footnotesize\setlength{\tabcolsep}{3pt}
  \begin{tabular}{@{}lccccc@{}}
    \toprule
    Dataset & Conv & Head & $\rho$ & Head Sav. & Total Red. \\
    \midrule
    CIFAR-10 & 51.3\% & 48.7\% & 48.7\% & 43.0\% & 20.9\% \\
    CIFAR-100 & 51.1\% & 48.9\% & 48.9\% & 42.4\% & 20.8\% \\
    Tiny-ImageNet & 81.9\% & 18.1\% & 18.1\% & 17.9\% & 3.2\% \\
    IN-1K (ResNet-18) & 99.9\% & 0.1\% & 0.06\% & --- & $\sim$0\% \\
    IN-1K (MobNetV2) & 99.2\% & 0.8\% & 0.85\% & --- & $\sim$0\% \\
    \midrule
    IN-1K (Backbone)\textsuperscript{$\dagger$} & 62\% & 38\% & 38\% & --- & --- \\
    \bottomrule
    \multicolumn{6}{@{}l}{\scriptsize\textsuperscript{$\dagger$}MoE in layer3/4; ``Head'' denotes MoE-routed FLOPs.}
  \end{tabular}
\end{table}

When the head constitutes fraction $\rho$ of total FLOPs, a sparse router that reduces head computation by factor $s$ saves $\rho \cdot s$ of total FLOPs. For very small $\rho$, even perfect routing yields negligible savings. Our experiments reveal a graded \emph{compute-leverage relationship} across six $\rho$ values (Table~\ref{tab:flops-decomp}): at high $\rho$ (CIFAR), routing yields both accuracy gains and meaningful FLOPs savings; at moderate $\rho$ (Tiny-ImageNet), accuracy gains but negligible FLOPs reduction; at $\rho \leq 1\%$ (ImageNet head-only), routing consistently hurts. MobileNet-V2 provides complementary evidence: it uses depthwise convolutions throughout yet still shows negative gaps, suggesting that $\rho$ is more predictive than backbone type. This observation is a CNN analogue of the capacity--compute trade-offs formalized by \citet{clark2022unified} for language MoE: routing helps when the routed computation is a non-trivial fraction of total compute. Backbone MoE (Section~\ref{subsec:backbone-moe}) recovers a positive gain on the same backbone, though confounds with routing regime preclude a purely causal interpretation (see Limitations). See supplementary Section~F for comparison against published efficient architectures.

\textbf{Adaptive Collapse at Low $\rho$.}
The qualitative routing behavior at low $\rho$ is consistent with this compute-leverage account. On ImageNet, ResNet-18 concentrates the majority of test-time routing on a single expert despite capacity constraints enforcing balanced usage per batch during training---the router learns a collapsed preference that emerges once capacity is removed at test time. This collapse is \emph{adaptive} rather than pathological: on MobileNet-V2, seeds with lower routing entropy achieve \emph{less} negative gaps, and the smallest expert size maintains near-perfect balance yet suffers the worst accuracy deficit. At low $\rho$, concentrating on one expert---effectively recovering a dense layer---appears to be the more efficient strategy, as distributing samples adds routing overhead without affecting meaningful computation.

We emphasize that the FLOPs analysis is \emph{analytical}: it quantifies the structural opportunity for routing to help, not practical inference speed. Table~\ref{tab:latency} reports wall-clock latency, on which our sparse MoE is the slowest method---a finding that warrants explanation rather than apology. The slowdown is a property of \textbf{how operations are scheduled, not how many are performed}. Routed inference activates only one of $E{=}8$ experts per token, so the FLOP count is in fact \emph{lower} than the matched dense head; the wall-clock cost arises elsewhere. Our implementation iterates over experts in a Python loop, invoking each expert's MLP via a separate sequence of CUDA kernel launches (gather, matmul, activation, scatter). Each launch incurs a fixed host-side dispatch overhead of order $10\,\mu$s, and across eight experts these constant-cost launches accumulate to dominate runtime; for small experts, the GPU spends most of its time idle between micro-kernels rather than computing.

Two observations confirm this attribution. First, the slowdown ratio \emph{collapses with model scale}: $145\times$ over dense on CIFAR-100, where the dense head itself is only 0.25 ms; $37\times$ on Tiny-ImageNet; and only $4.1\times$ on ImageNet-1K, where the 11.3 ms ResNet-18 backbone dwarfs the routing overhead. The trajectory is consistent with a fixed per-step launch cost rather than a workload-proportional one, and it implies favorable scaling: deeper backbones, higher input resolutions, or larger batch sizes further amortize the constant dispatch overhead. The settings in which routing structurally helps (high $\rho$ at large scale) are precisely the settings in which the unfused-implementation penalty matters least. Second, the contrast with the other routing methods in Table~\ref{tab:latency} is a contrast in \emph{dispatch expression} on identical hardware: Soft MoE reduces dispatch to a single batched matmul (0.63 ms on CIFAR-100), and Expert Choice's expert-side grouping (7.39 ms) is more GEMM-friendly than our per-expert sequential calls---a roughly $60\times$ Soft-MoE-vs-ours gap on the same model and the same GPU. Fused MoE kernels~\cite{gale2023megablocks,hwang2023tutel} have demonstrated that top-$k$ routing can approach dense throughput when per-expert dispatch is fused into a single grouped-GEMM call; we report our latency as an upper bound on production cost. The accuracy gain under matched training stands.

\begin{table}[t]
  \centering
  \caption{Inference latency (ms) and throughput (kimg/s) at batch size 128. %
  Measured on RTX 5090, CUDA sync (50 warmup + 200 timed passes).}
  \label{tab:latency}
  \scriptsize
  \setlength{\tabcolsep}{3pt}
  \begin{tabular}{@{}lrrrrrr@{}}
    \toprule
    & \multicolumn{2}{c}{C-100} & \multicolumn{2}{c}{Tiny-IN} & \multicolumn{2}{c}{IN-1K} \\
    \cmidrule(lr){2-3} \cmidrule(lr){4-5} \cmidrule(lr){6-7}
    Method & ms & k/s & ms & k/s & ms & k/s \\
    \midrule
    Dense          & 0.25  & 519.3 & 1.01  & 126.2 & 11.3  & 11.3 \\
    MoE $k{=}1$   & 36.38 &   3.5 & 37.44 &   3.4 & 46.8  &  2.7 \\
    Soft MoE       & 0.63  & 204.2 & 1.31  &  97.6 & 11.4  & 11.2 \\
    Expert Choice  & 7.39  &  17.3 & 7.62  &  16.8 & 17.6  &  7.3 \\
    \bottomrule
  \end{tabular}
\end{table}

\subsection{Cross-Method Synthesis}

Table~\ref{tab:cross-dataset-methods} consolidates the cross-method comparison across all four datasets. Three patterns emerge: (1)~token-mixing Soft MoE collapses universally, worsening with task complexity; (2)~per-sample soft gating (Section~\ref{subsec:persample-soft}) eliminates this collapse on every dataset, recovering to near-dense levels but never consistently surpassing them; and (3)~sparse MoE with depthwise backbones achieves positive gaps when $\rho$ is sufficiently high.

\textbf{Diagnosing the Soft MoE failure.} We probe the dispatch mechanism with a per-sample variant that softmaxes over experts independently per image (Table~\ref{tab:c2-persample}). On CIFAR-100, per-sample dispatch rescues the catastrophe ($-36\% \to +1.69\%$), establishing batch-axis dispatch as the dominant failure mode in our per-sample CNN setting. On Tiny-ImageNet ($-1.13\%$) and ImageNet ($-0.66\%$), per-sample dispatch is no longer catastrophic but does not reach parity, indicating a residual scale-dependent factor that we do not isolate (one possibility is that weighted expert ensembling without hard routing cannot fully exploit pretrained features at scale).

\begin{table}[t]
  \centering
  \caption{Per-sample soft gating diagnostic (5-seed mean$\pm$s.d., peak validation accuracy). Dispatch is the dominant Soft MoE failure mode but not the only one.}
  \label{tab:c2-persample}
  \footnotesize\setlength{\tabcolsep}{3pt}
  \begin{tabular}{@{}lccccc@{}}
    \toprule
    Dataset & Dense & Soft MoE & Gap & sd & $t$ \\
    \midrule
    CIFAR-100      & 57.53 & 59.22 & $\mathbf{+1.69}$ & 0.31 & $+12.02$ \\
    Tiny-ImageNet  & 45.83 & 44.70 & $-1.13$ & 0.97 & $-2.60$ \\
    ImageNet-1K    & 70.33 & 69.67 & $-0.66$ & 0.08 & $-19.45$ \\
    \bottomrule
  \end{tabular}
\end{table}

\begin{table}[t]
  \centering
  \caption{Cross-method accuracy gaps (\%) across all datasets. Best non-dense result \textbf{bolded}. %
  \textsuperscript{$\star$}Per-sample soft gating (5-seed mean for C-100/Tiny-IN/IN-1K).}
  \label{tab:cross-dataset-methods}
  \footnotesize\setlength{\tabcolsep}{3pt}
  \begin{tabular}{@{}lccccc@{}}
    \toprule
    Method & C-10 & C-100 & Tiny-IN & IN-1K & Trend \\
    \midrule
    Soft MoE (token-mix) & $-13.3$ & $-36.4$ & $-42.9$ & $-29.3$ & $\downarrow$ \\
    Soft MoE (per-sample)\textsuperscript{$\star$} & --- & $+1.7$ & $-1.1$ & $-0.7$ & $\approx$ \\
    Expert Choice & $-1.2$ & $-6.8$ & $-4.2$ & $-29.0$ & $\downarrow$ \\
    Sparse (std.) & $-1.4$ & $-4.0$ & $-0.3$ & $-1.3$ & $\approx$ \\
    \midrule
    Ours (DW $w{=}0.72$) & $\mathbf{+1.3}$ & $\mathbf{+3.0}$ & $+3.4$ & --- & $\uparrow$ \\
    Ours (wider DW) & $+0.5$ & $+1.4$ & $\mathbf{+4.0}$ & --- & $\uparrow$ \\
    \midrule
    \multicolumn{6}{@{}l}{\emph{ImageNet-1K (pretrained, head-only, $\rho < 1\%$):}} \\
    Ours (ResNet-18) & --- & --- & --- & $-1.3$ & --- \\
    Ours (MobNetV2) & --- & --- & --- & $-1.9$ & --- \\
    \midrule
    \multicolumn{6}{@{}l}{\emph{ImageNet-1K (backbone MoE, $\rho \approx 38\%$):}} \\
    Ours (layer3/4) & --- & --- & --- & $\mathbf{+1.2}$ & $\uparrow$ \\
    \bottomrule
  \end{tabular}
\end{table}

Only our method shows an upward trend across the three smaller-scale datasets; head-only ImageNet results confirm the negative-$\rho$ boundary, while backbone MoE recovers a positive gap on the same dataset. On standard backbones, all methods show flat or declining gaps, consistent with the contribution being the joint recipe rather than routing alone. Widening the DW backbone resolves the absolute accuracy gap: DW+MoE exceeds standard dense on all three datasets, confirming that narrow and wide configurations represent different operating points on the same principle.

\section{Conclusion}
\label{sec:conclusion}

We presented an empirical study of when sparse MoE routing helps in vision classification, identifying a \emph{compute-leverage pattern}: positive accuracy gaps require a substantial fraction $\rho$ of total FLOPs to be routed. A controlled hidden-size sweep on CIFAR-10 (Section~\ref{subsec:rho-sweep}, Table~\ref{tab:c1-rho-sweep}) yields a monotonic gap--$\rho$ relationship and both predicted sign reversals, ruling out backbone family as the active variable. At ImageNet scale, this fraction-condition becomes necessary but not sufficient: a complementary $k$-ablation (Section~\ref{subsec:backbone-moe}) holds backbone, routed fraction, and pretrained initialization fixed and varies only the top-$k$ selection, reversing the gap from $+1.17\%$ to $-2.08\%$ across all five seeds. The MobileNet-V2 control reinforces this picture: depthwise convolutions alone are insufficient when the routed fraction is small. Across our experiments, then, a substantial $\rho$ is necessary at every scale, and at ImageNet scale multi-expert routing ($k \geq 2$) is additionally required; neither alone suffices.

Within the favorable $\rho$ regime, MoE--dense gaps trend upward with task complexity across three benchmarks, though these comparisons span different backbone widths and evaluation protocols. Wider depthwise backbones reach competitive absolute accuracy, exceeding standard dense baselines on all three smaller-scale benchmarks. On standard backbones, both token-mixing Soft MoE and Expert Choice underperform the dense baseline in our per-sample CNN setting. A diagnostic per-sample variant that softmaxes over experts rather than over the batch rescues CIFAR-100 above the dense baseline, identifying batch-axis dispatch as the dominant Soft MoE failure mode, but leaves Tiny-ImageNet and ImageNet-1K negative---a residual, scale-dependent factor we do not isolate. We use hard capacity constraints to guarantee per-batch expert usage during training, and temperature scheduling acts as a complementary lever for late-stage stability.

\textbf{Limitations.} An additive utility bias explored during development was found to be operationally negligible in the CIFAR experiments: code inspection confirmed it did not materially affect routing decisions, and a CIFAR-100 ablation showed no difference when it was disabled ($p{=}0.87$; supplementary Section~I). It is not claimed as a contribution. The evolutionary search uses CIFAR-10's test split for fitness; a held-out validation split would improve protocol cleanliness. Training recipes use standard optimizers and augmentations rather than modern techniques (AdamW, CutMix); relative comparisons remain valid under matched training, but absolute accuracy is not competitive with stronger recipes. Our backbone-MoE setup uses pretrained initialization with $k{=}2$ routing, while the head-only protocol uses random initialization with $k{=}1$; the $k$-ablation in Section~\ref{subsec:backbone-moe} bridges these on a single architecture, but a fully unified protocol across all four datasets remains open. The CIFAR-10 $\rho$-sweep (Section~\ref{subsec:rho-sweep}) varies head capacity and the routed fraction jointly; a strict isolation experiment with FLOP-matched dense controls is left to future work. Wall-clock latency in our unfused implementation is dominated by per-expert kernel launches rather than expert arithmetic (Section~\ref{subsec:flops-decomp}); the gap to dense throughput is closed in principle by fused MoE kernels~\cite{gale2023megablocks,hwang2023tutel} but not yet integrated here.

\textbf{Future Work.} Promising directions include extending backbone MoE to deeper architectures and larger scales, integrating fused dispatch kernels to close the implementation-side latency gap, knowledge distillation~\cite{hinton2015distilling} from MoE to dense students, and validation on fine-grained domains such as iNaturalist.

\appendix
\section{Additional Routing Visualizations}

\subsection{Routing Heatmaps}
\begin{figure}[ht]
  \centering
  \includegraphics[width=0.95\linewidth]{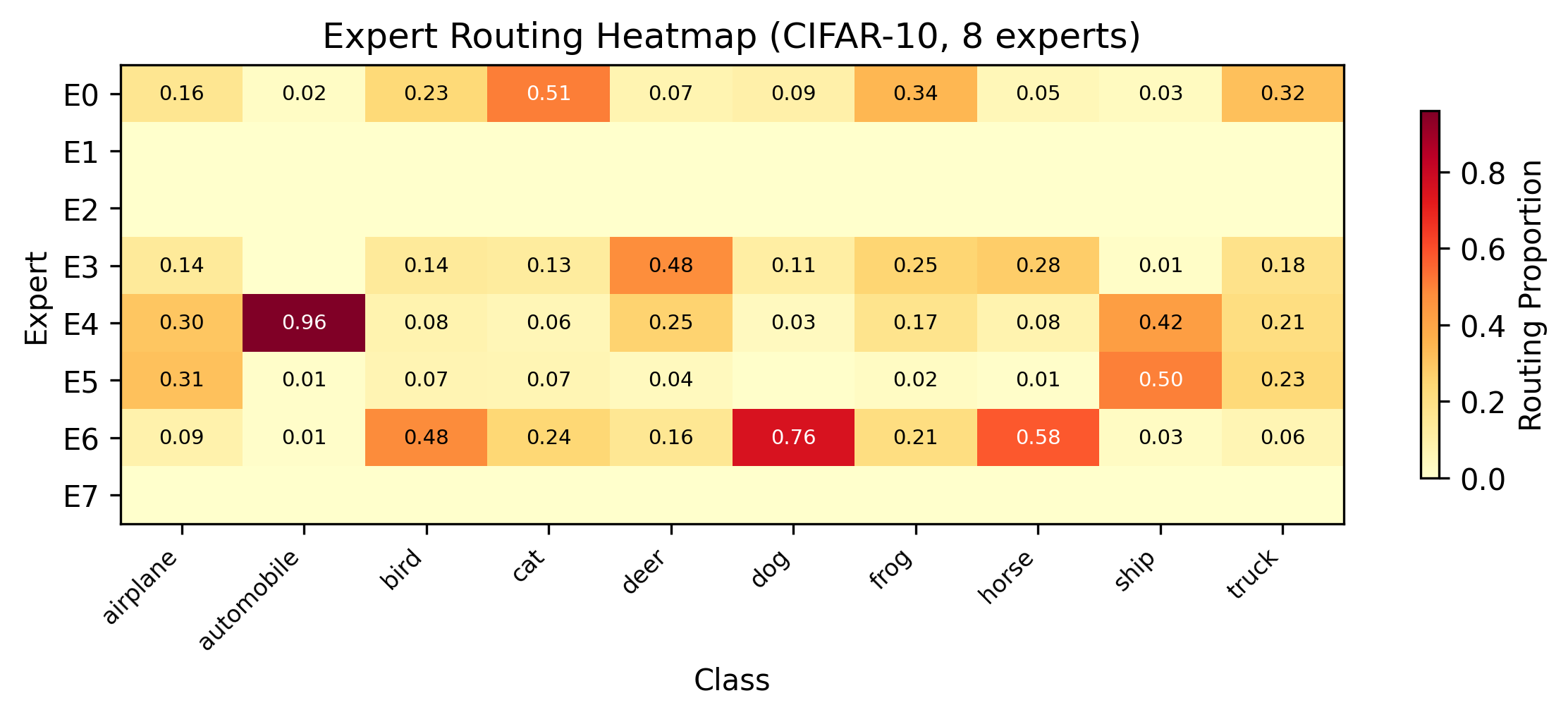}
  \caption{Expert routing heatmap, CIFAR-10 test set (3-block+BN DW $w{=}2.0$, seed 2024). Five of eight experts show clear class specialization.}
  \label{fig:supp-routing-heatmap-cifar10}
\end{figure}

The CIFAR-10 routing heatmap (Fig.~\ref{fig:supp-routing-heatmap-cifar10}) shows expert specialization across 10 classes using the 3-block+BN DW $w{=}2.0$ backbone. Five of eight experts show clear class specialization, with Expert 4 dominating automobiles at 0.96. The 10-class setting produces sharper specialization than the CIFAR-100 heatmap in Fig.~\ref{fig:routing-heatmap}, where each expert covers more classes.

\subsection{$t$-SNE of CIFAR-100 Routing}
\begin{figure}[ht]
  \centering
  \includegraphics[width=\linewidth]{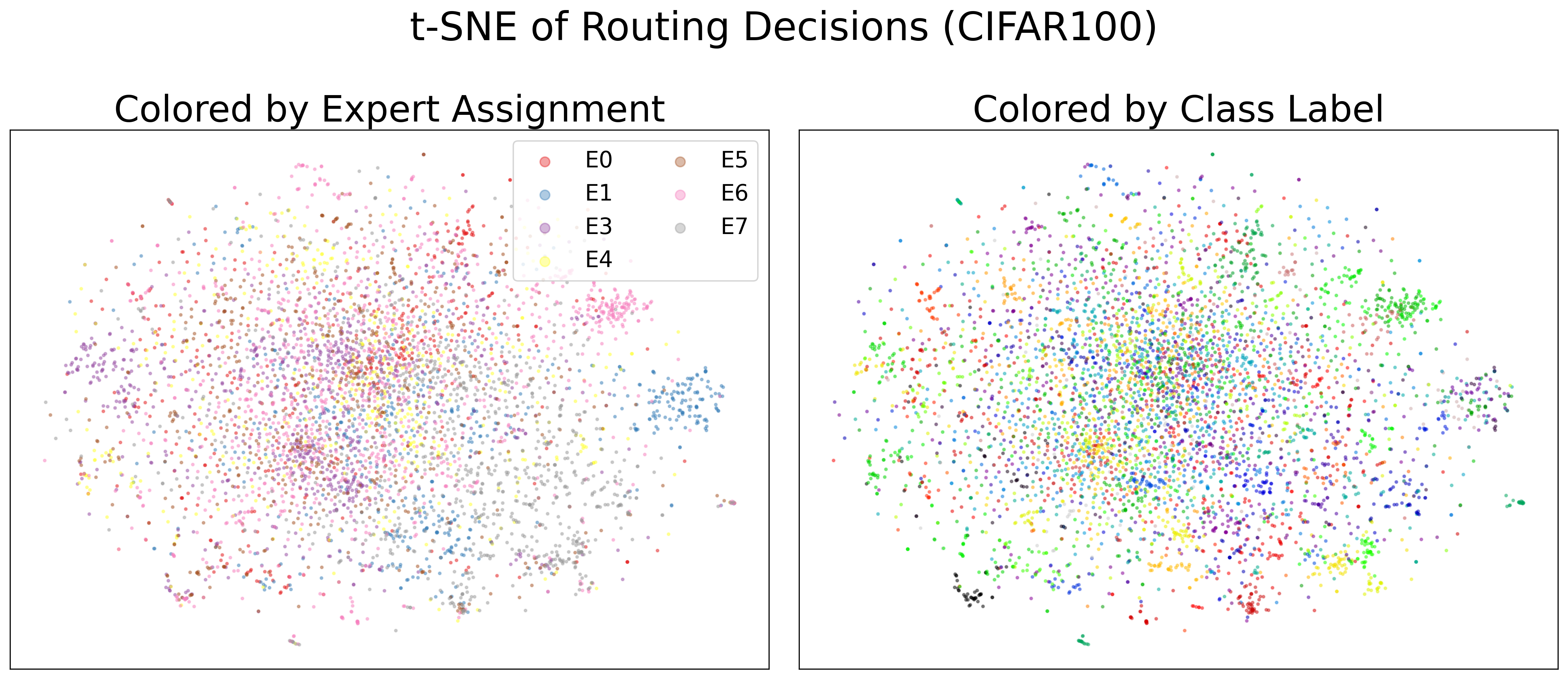}
  \caption{$t$-SNE of CIFAR-100 routing (3-block+BN DW $w{=}2.0$, seed 123). Left: by expert assignment. Right: by class label.}
  \label{fig:supp-tsne-routing}
\end{figure}

Fig.~\ref{fig:supp-tsne-routing} shows $t$-SNE embeddings of the CIFAR-100 test set colored by expert assignment (left) and by class label (right). The two panels show visually consistent structure, but the expert assignments are not independent of the class-cluster geometry and we treat the alignment as qualitative rather than as a quantitative claim of semantic specialization.

\section{Training Dynamics}

\begin{figure}[ht]
  \centering
  \includegraphics[width=0.95\linewidth]{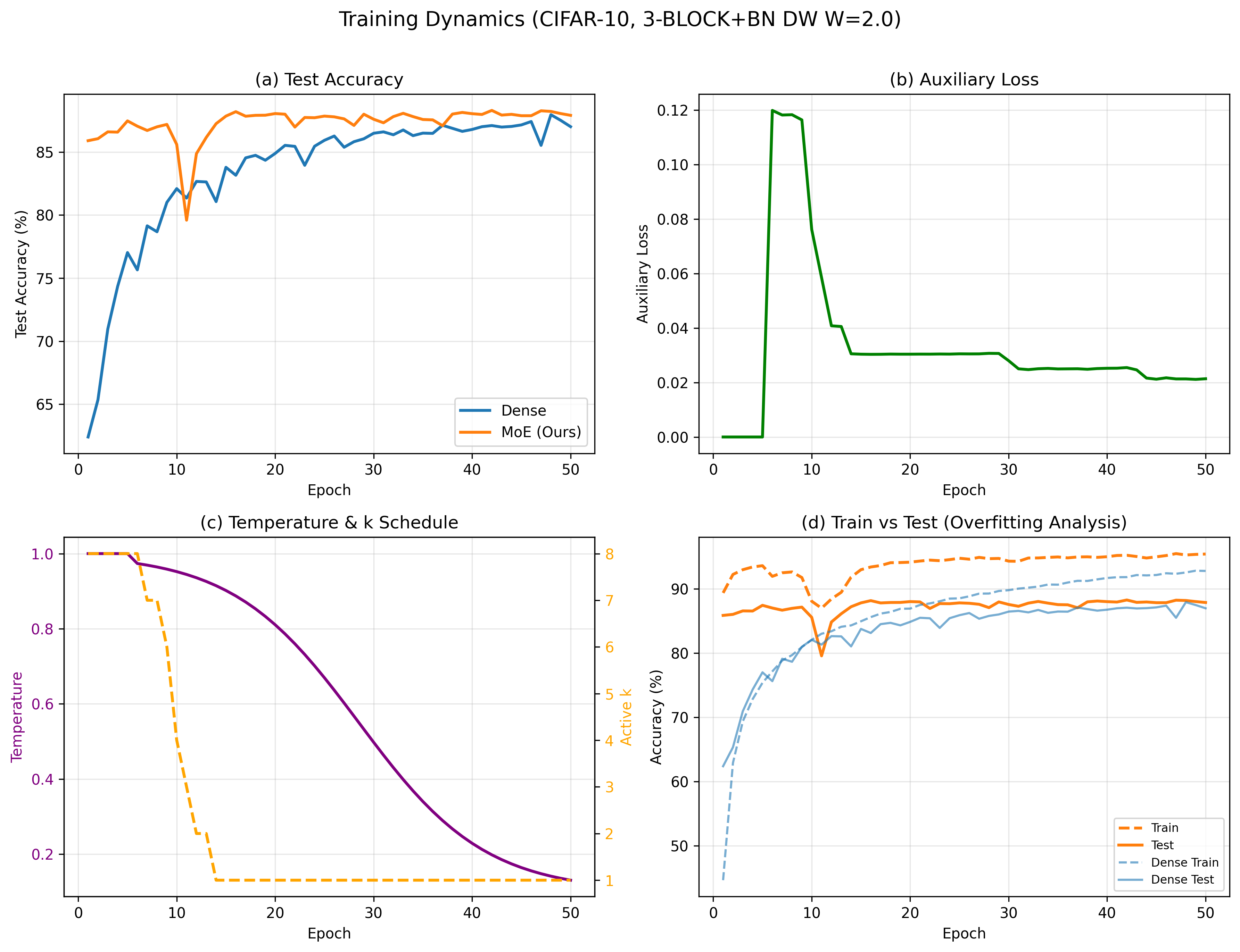}
  \caption{Training dynamics on CIFAR-10 (3-block+BN DW $w{=}2.0$, seed 2024). MoE surpasses dense after epoch 10.}
  \label{fig:supp-dynamics-cifar10}
\end{figure}

\begin{figure}[ht]
  \centering
  \includegraphics[width=0.95\linewidth]{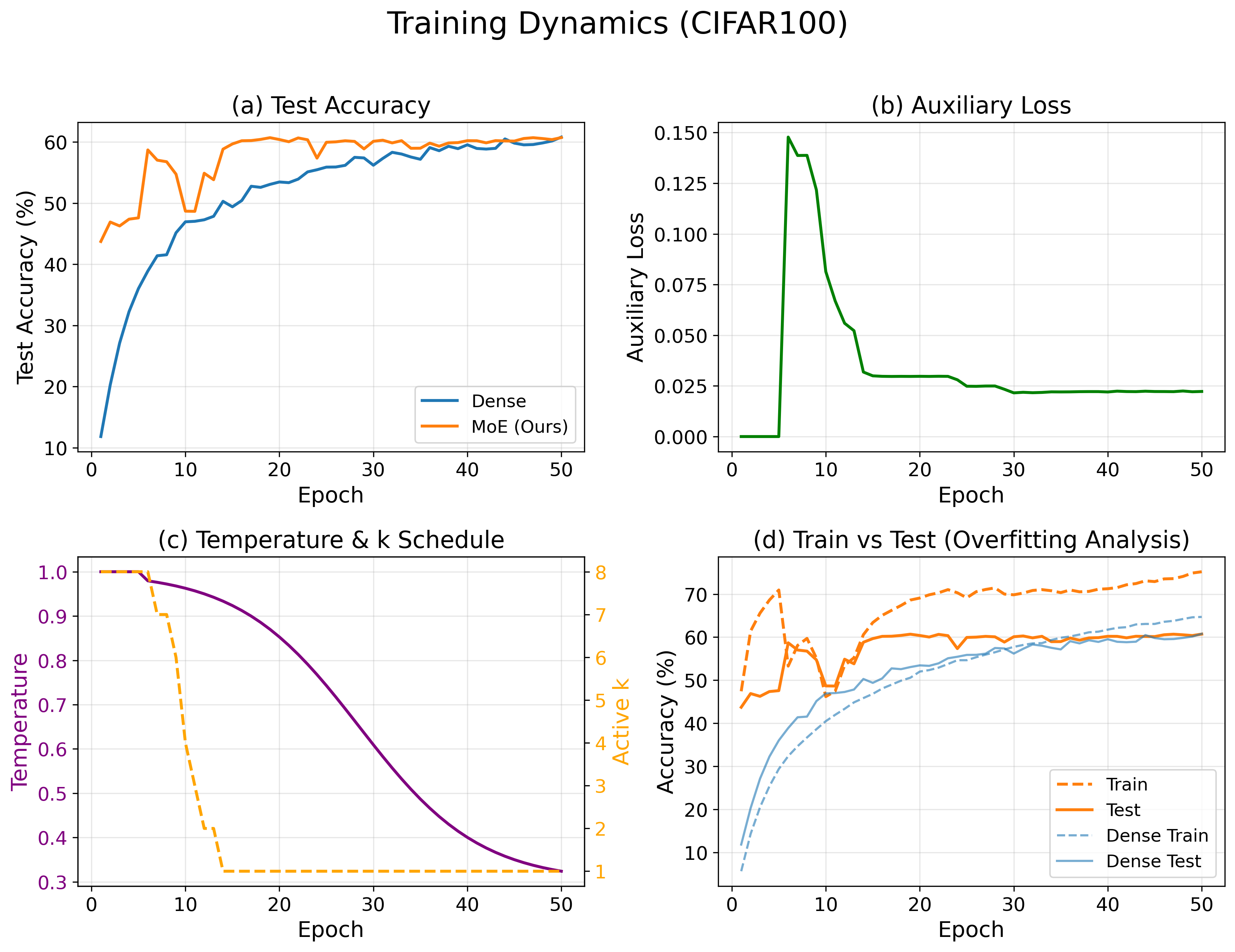}
  \caption{Training dynamics on CIFAR-100 (3-block+BN DW $w{=}2.0$, seed 123). (a)~Accuracy gap. (b)~Auxiliary loss. (c)~Schedule. (d)~Train/test accuracy.}
  \label{fig:supp-dynamics-cifar100}
\end{figure}

\begin{figure}[ht]
  \centering
  \includegraphics[width=0.95\linewidth]{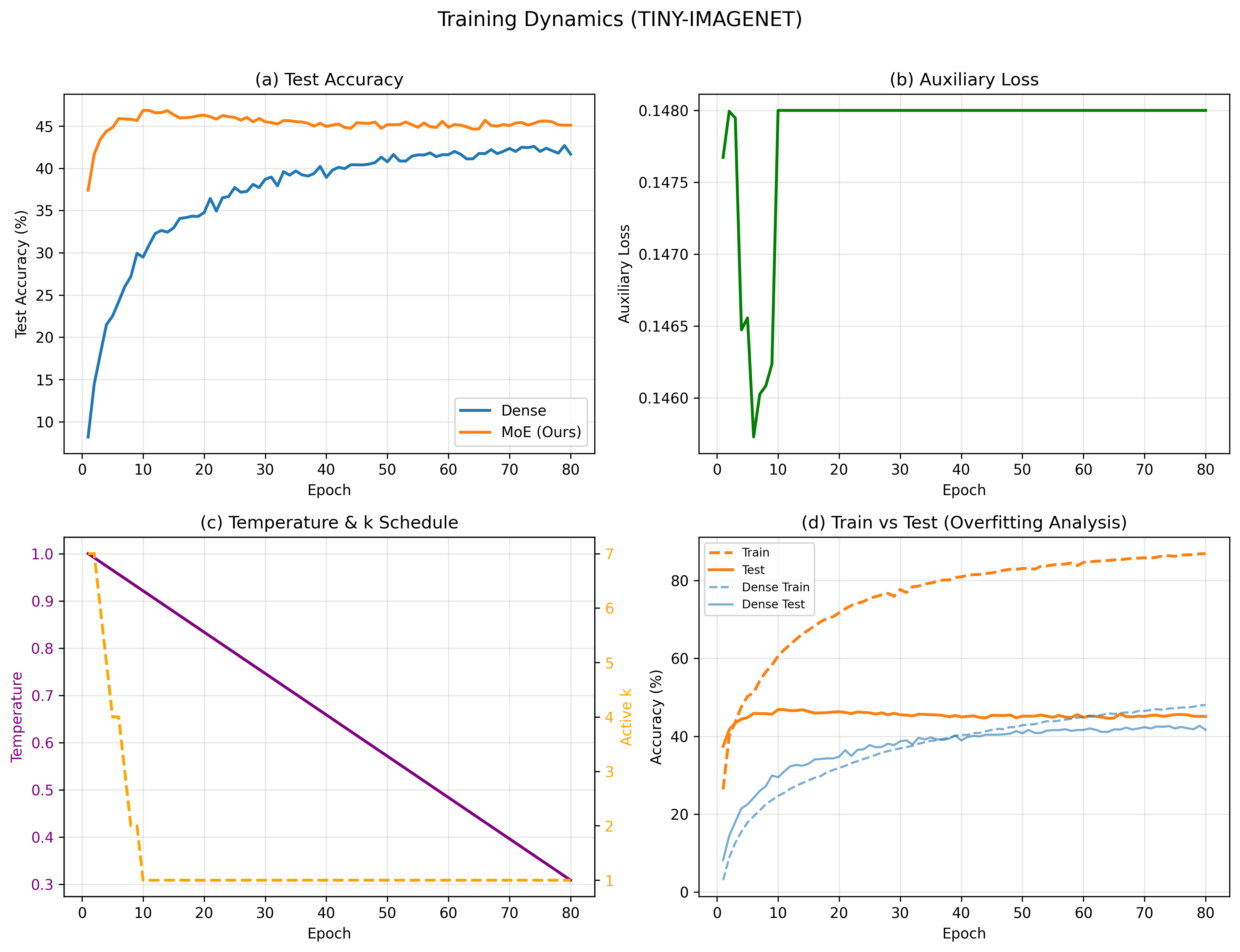}
  \caption{Training dynamics on Tiny-ImageNet (DW $w{=}1.2$, seed 42). MoE outperforms dense by $+4.2\%$ at its best epoch; routing entropy decreases smoothly from $\sim$2.0 to $\sim$1.5 as temperature anneals.}
  \label{fig:supp-dynamics-tiny}
\end{figure}

Figures~\ref{fig:supp-dynamics-cifar10}, \ref{fig:supp-dynamics-cifar100}, and \ref{fig:supp-dynamics-tiny} show training dynamics for the three smaller-scale datasets. Across all three, expert usage stabilizes early in training, MoE surpasses the dense baseline by mid-training, and the gap is maintained with comparable train--test generalization, suggesting improved representation rather than overfitting.

\section{Temperature Schedule Ablation Details}

Table~\ref{tab:supp-temperature-ablations} presents temperature schedule ablations on the standard CIFAR-10 backbone. The configuration with linear annealing to $\tau_{\min}{=}0.1$ achieves $+0.92\%$ peak over the dense baseline but collapses in late stages (final accuracy 80.49\%). Sigmoid schedules avoid collapse entirely ($+0.03\%$ gap, stable). Setting $\tau_{\min}{=}0.3$ (as used on Tiny-ImageNet and ImageNet) prevents collapse with linear schedules. These results motivated the per-dataset schedule choices described in Section~\ref{subsec:setup}. The configurations all include the additive utility bias, which is itself ablated separately in Section~\ref{sec:supp-utility-ablation}.

\begin{table}[ht]
  \centering
  \caption{Temperature schedule ablations on the standard CIFAR-10 backbone. All rows include the additive utility bias; the utility ablation is in Section~\ref{sec:supp-utility-ablation}.}
  \label{tab:supp-temperature-ablations}
  \small
  \begin{tabular}{@{}lcccc@{}}
    \toprule
    Config & Dense & Peak & Final & $\Delta$FLOPs \\
    \midrule
    $\tau$: 1.0$\to$0.1 (linear) & 83.34 & \textbf{84.26} & 80.49 & $-$9.4\% \\
    $\tau$: 1.0$\to$0.5 (linear) & 84.00 & 81.20 & 76.50 & $-$36.2\% \\
    $\tau$: 1.0$\to$0.13 (sigmoid, $\kappa{=}7$) & 83.38 & 83.41 & 83.41 & $-$8.9\% \\
    \bottomrule
  \end{tabular}
\end{table}

\section{Hyperparameter Search Details}
\label{sec:supp-search}

Table~\ref{tab:supp-hparam-bounds} lists the four hyperparameters varied by the evolutionary search on CIFAR-10 (Section~\ref{sec:evo-search}). Ranges are narrow by design: the search protocol is a lightweight tuning loop rather than a NAS contribution. The three routing hyperparameters ($c$, $\lambda_{\text{lb}}$, $\lambda_{\text{ent}}$) transfer to all three evaluation datasets without retuning; only training details ($h$, $\tau_{\min}$, warmup epochs) are adjusted per dataset.

\begin{table}[ht]
  \centering
  \caption{Evolutionary search ranges and optimized values. A utility weight $\lambda_u$ was also searched ($[0.08, 0.15]$, best $0.087$) but found operationally negligible; see development note in Section~\ref{sec:methodology}.}
  \label{tab:supp-hparam-bounds}
  \small
  \begin{tabular}{@{}lccc@{}}
    \toprule
    Parameter & Range & Mut.\ $\sigma$ & Best \\
    \midrule
    Width scale $w$ & $[0.65, 0.85]$ & 0.020 & 0.717 \\
    Capacity $c$ & $[1.05, 1.20]$ & 0.015 & 1.064 \\
    Load-bal.\ $\lambda_{\text{lb}}$ & $[0.01, 0.02]$ & 0.001 & 0.019 \\
    Entropy $\lambda_{\text{ent}}$ & $[0.015, 0.030]$ & 0.002 & 0.024 \\
    \bottomrule
  \end{tabular}
\end{table}

\section{Per-Sample Soft Gating Derivation}
\label{sec:supp-persample}

Standard Soft MoE~\cite{puigcerver2024soft} computes dispatch weights via:
\begin{equation}
  D_{ij} = \frac{\exp(\phi_{ij})}{\sum_{i'=1}^{B} \exp(\phi_{i'j})}, \quad \phi_{ij} = x_i^\top s_j,
  \label{eq:supp-dispatch}
\end{equation}
where $x_i$ is the $i$-th input token and $s_j$ the $j$-th slot embedding. The softmax is over the batch (token) dimension ($i$). Each slot $j$ receives a weighted combination $\tilde{x}_j = \sum_i D_{ij} x_i$ of all input tokens.

In Vision Transformers, tokens are semantically related patches from the same image, so this averaging produces a meaningful aggregate. In our per-sample CNN classification setting, each ``token'' is an \emph{independent image}. With batch size $B$ and $E$ experts (each with one slot), the dispatch weights become:
\begin{equation}
  D_{ij} \approx \frac{1}{B} \quad \text{for large } B,
  \label{eq:supp-dispatch-approx}
\end{equation}
because the softmax over $B$ unrelated inputs produces near-uniform weights. Each expert slot thus receives an approximately uniform average of unrelated images, destroying discriminative information.

Our \emph{per-sample soft gating} (Eqs.~\ref{eq:persample-gate}--\ref{eq:persample-combine}) addresses this by applying the softmax over experts instead of tokens:
\begin{equation}
  w_i^{(b)} = \frac{\exp(h_b^\top s_i / \tau)}{\sum_{j=1}^{E} \exp(h_b^\top s_j / \tau)}, \quad o_b = \sum_{i=1}^{E} w_i^{(b)} \cdot \text{Expert}_i(h_b).
\end{equation}
This preserves per-sample discriminative information: each sample is independently routed through all experts with sample-specific combination weights, eliminating the cross-sample averaging. The resulting model is equivalent to a soft attention over expert outputs, weighted by input--slot compatibility.

\section{Published Efficiency Baselines}
\label{sec:supp-efficiency}

Table~\ref{tab:supp-efficiency} contextualizes our best models against published baselines across all four datasets. Our study uses lightweight backbones to isolate MoE routing effects under controlled conditions; the comparison clarifies the accuracy--efficiency operating point relative to standard architectures trained with stronger recipes.

\begin{table}[ht]
  \centering
  \caption{Absolute accuracy of our best models vs.\ published baselines. Published numbers use standard training recipes (AdamW, CutMix, 200+ epochs); our models use basic recipes (Adam/SGD, 50--80 epochs) to control for training effects. $^\dagger$Approximate published numbers.}
  \label{tab:supp-efficiency}
  \small
  \begin{tabular}{@{}llcc@{}}
    \toprule
    Dataset & Model & Acc (\%) & FLOPs \\
    \midrule
    \multirow{4}{*}{CIFAR-10}
    & ResNet-18 (standard recipe)$^\dagger$ & $\sim$95 & 556M \\
    & MobileNet-V2$^\dagger$ & $\sim$94 & 300M \\
    & Ours (DW $w{=}2.0$ + MoE) & 88.41 & $\sim$24M \\
    & Ours (DW $w{=}0.72$ + MoE) & 78.62 & $\sim$3M \\
    \midrule
    \multirow{3}{*}{CIFAR-100}
    & ResNet-18 (standard recipe)$^\dagger$ & $\sim$78 & 556M \\
    & MobileNet-V2$^\dagger$ & $\sim$78 & 300M \\
    & Ours (DW $w{=}2.0$ + MoE) & 61.29 & $\sim$24M \\
    \midrule
    \multirow{3}{*}{Tiny-IN}
    & ResNet-18 (standard recipe)$^\dagger$ & $\sim$66 & 1.8G \\
    & MobileNet-V2$^\dagger$ & $\sim$64 & 300M \\
    & Ours (DW $w{=}1.2$ + MoE) & 46.35 & $\sim$90M \\
    \midrule
    \multirow{3}{*}{ImageNet}
    & ResNet-18 (torchvision) & 69.76 & 1.8G \\
    & MobileNet-V2 (torchvision) & 72.15 & 300M \\
    & Our Dense (MobNetV2 fine-tuned) & 72.26 & 300M \\
    \bottomrule
  \end{tabular}
\end{table}

Two observations emerge. First, the absolute accuracy gap between our lightweight models and published baselines is substantial on CIFAR/Tiny-ImageNet (20--30 percentage points), reflecting the deliberately constrained backbone design and basic training recipe rather than a deficiency of the MoE mechanism. Second, on ImageNet where we use standard pretrained backbones, our dense baselines match published numbers ($72.26\%$ vs.\ torchvision's $72.15\%$ for MobileNet-V2), confirming that the negative MoE gaps on ImageNet are attributable to the low $\rho$ regime rather than implementation issues. The key contribution of our study is the \emph{relative} MoE--dense improvement under matched conditions, which isolates the effect of routing from confounding training variables.

\section{Tier-C Strengthening Experiments: Per-Seed Results}
\label{sec:supp-tierc}

This section reports per-seed accuracies for the three Tier-C experiments aggregated in Sections~\ref{subsec:rho-sweep}, \ref{subsec:backbone-moe}, and~\ref{sec:analysis}: the controlled $\rho$-sweep on CIFAR-10, the per-sample soft gating diagnostic across three datasets, and the backbone-MoE $k{=}1$ ablation on ImageNet-1K. All 50 source JSONs are listed in \texttt{phase2\_moe\_backbone/TIER\_C\_FINDINGS.md} of the code repository (\url{https://github.com/libophd/sparse-moe-vision-rho}); aggregation is performed by \texttt{scripts/cluster/aggregate\_summary.py}.

\subsection{C1 --- CIFAR-10 $\rho$-sweep (per-seed)}

Each row reports final-epoch test accuracy for one (config, seed) pair. Six configurations $\times$ five seeds $=$ 30 runs.

\begin{table}[ht]
  \centering
  \caption{Per-seed final-epoch test accuracy (\%) for the CIFAR-10 $\rho$-sweep.}
  \label{tab:supp-c1-perseed}
  \footnotesize\setlength{\tabcolsep}{3pt}
  \begin{tabular}{@{}lrrrrr@{}}
    \toprule
    Config & Seed & Dense & MoE & Gap \\
    \midrule
    Std.\ $h{=}128$  & 42   & 84.56 & 81.84 & $-2.72$ \\
    Std.\ $h{=}128$  & 123  & 84.43 & 82.23 & $-2.20$ \\
    Std.\ $h{=}128$  & 456  & 83.66 & 81.40 & $-2.26$ \\
    Std.\ $h{=}128$  & 777  & 84.58 & 82.29 & $-2.29$ \\
    Std.\ $h{=}128$  & 2025 & 84.48 & 83.06 & $-1.42$ \\
    Std.\ $h{=}512$  & 42   & 83.77 & 82.80 & $-0.97$ \\
    Std.\ $h{=}512$  & 123  & 83.80 & 83.56 & $-0.24$ \\
    Std.\ $h{=}512$  & 456  & 83.38 & 83.12 & $-0.26$ \\
    Std.\ $h{=}512$  & 777  & 83.42 & 82.82 & $-0.60$ \\
    Std.\ $h{=}512$  & 2025 & 83.19 & 83.89 & $+0.70$ \\
    Std.\ $h{=}2048$ & 42   & 82.90 & 85.21 & $+2.31$ \\
    Std.\ $h{=}2048$ & 123  & 83.42 & 85.39 & $+1.97$ \\
    Std.\ $h{=}2048$ & 456  & 83.26 & 82.85 & $-0.41$ \\
    Std.\ $h{=}2048$ & 777  & 83.03 & 83.96 & $+0.93$ \\
    Std.\ $h{=}2048$ & 2025 & 82.61 & 83.15 & $+0.54$ \\
    DW $h{=}128$     & 42   & 78.01 & 78.43 & $+0.42$ \\
    DW $h{=}128$     & 123  & 76.99 & 77.26 & $+0.27$ \\
    DW $h{=}128$     & 456  & 75.54 & 75.95 & $+0.41$ \\
    DW $h{=}128$     & 777  & 78.00 & 74.75 & $-3.25$ \\
    DW $h{=}128$     & 2025 & 76.22 & 72.00 & $-4.22$ \\
    DW $h{=}512$     & 42   & 76.06 & 78.41 & $+2.35$ \\
    DW $h{=}512$     & 123  & 76.56 & 76.72 & $+0.16$ \\
    DW $h{=}512$     & 456  & 75.83 & 79.94 & $+4.11$ \\
    DW $h{=}512$     & 777  & 76.29 & 79.73 & $+3.44$ \\
    DW $h{=}512$     & 2025 & 76.53 & 75.99 & $-0.54$ \\
    DW $h{=}2048$    & 42   & 74.61 & 81.04 & $+6.43$ \\
    DW $h{=}2048$    & 123  & 75.81 & 81.12 & $+5.31$ \\
    DW $h{=}2048$    & 456  & 75.56 & 81.98 & $+6.42$ \\
    DW $h{=}2048$    & 777  & 77.36 & 80.25 & $+2.89$ \\
    DW $h{=}2048$    & 2025 & 76.15 & 79.06 & $+2.91$ \\
    \bottomrule
  \end{tabular}
\end{table}

\subsection{C2 --- Per-sample soft gating (per-seed)}

Five seeds per dataset; peak validation accuracy.

\begin{table}[ht]
  \centering
  \caption{Per-seed peak validation accuracy (\%) for the per-sample soft gating diagnostic.}
  \label{tab:supp-c2-perseed}
  \footnotesize\setlength{\tabcolsep}{3pt}
  \begin{tabular}{@{}lrrrr@{}}
    \toprule
    Dataset & Seed & Dense & Soft MoE & Gap \\
    \midrule
    CIFAR-100      & 42   & 58.16 & 59.58 & $+1.42$ \\
    CIFAR-100      & 123  & 57.99 & 59.35 & $+1.36$ \\
    CIFAR-100      & 456  & 57.65 & 59.41 & $+1.76$ \\
    CIFAR-100      & 777  & 56.75 & 58.89 & $+2.14$ \\
    CIFAR-100      & 2025 & 57.08 & 58.85 & $+1.77$ \\
    Tiny-ImageNet  & 42   & 45.46 & 43.55 & $-1.91$ \\
    Tiny-ImageNet  & 123  & 46.40 & 44.72 & $-1.68$ \\
    Tiny-ImageNet  & 456  & 45.64 & 45.92 & $+0.28$ \\
    Tiny-ImageNet  & 777  & 45.96 & 44.15 & $-1.81$ \\
    Tiny-ImageNet  & 2025 & 45.67 & 45.16 & $-0.51$ \\
    ImageNet-1K    & 42   & 70.36 & 69.72 & $-0.64$ \\
    ImageNet-1K    & 123  & 70.32 & 69.68 & $-0.64$ \\
    ImageNet-1K    & 456  & 70.32 & 69.72 & $-0.60$ \\
    ImageNet-1K    & 777  & 70.34 & 69.69 & $-0.65$ \\
    ImageNet-1K    & 2025 & 70.33 & 69.53 & $-0.80$ \\
    \bottomrule
  \end{tabular}
\end{table}

\subsection{C3 --- Backbone-MoE $k{=}1$ ablation (per-seed)}

Five seeds, peak top-1 validation accuracy on ImageNet-1K. Same backbone, same $\rho \approx 38\%$, same pretrained initialization as the headline $k{=}2$ result; only the routing top-$k$ changes.

\begin{table}[ht]
  \centering
  \caption{Per-seed peak top-1 (\%) for the backbone-MoE $k{=}1$ ablation on ImageNet-1K.}
  \label{tab:supp-c3-perseed}
  \small
  \begin{tabular}{@{}rrrr@{}}
    \toprule
    Seed & Dense & MoE $k{=}1$ & Gap \\
    \midrule
    42   & 69.99 & 68.06 & $-1.93$ \\
    123  & 70.10 & 67.96 & $-2.14$ \\
    456  & 70.04 & 67.91 & $-2.13$ \\
    777  & 70.03 & 67.76 & $-2.27$ \\
    2025 & 69.97 & 68.04 & $-1.93$ \\
    \midrule
    Mean$\pm$s.d. & $70.03\pm0.05$ & $67.95\pm0.12$ & $\mathbf{-2.08\pm0.15}$ \\
    \bottomrule
  \end{tabular}
\end{table}

\subsection{Backbone-MoE $k{=}2$ headline (per-seed)}

The 5-seed per-seed accuracies for the headline backbone-MoE result on ImageNet-1K (Section~\ref{subsec:backbone-moe}). Companion to the $k{=}1$ ablation in the previous subsection.

\begin{table}[ht]
  \centering
  \caption{Per-seed peak top-1 (\%) for backbone-MoE $k{=}2$ on ImageNet-1K (ResNet-18 layer3/4 replaced with MoEConv2d, $E{=}8$, $\rho \approx 38\%$).}
  \label{tab:supp-backbone-moe-k2}
  \small
  \begin{tabular}{@{}rrrr@{}}
    \toprule
    Seed & Dense & MoE $k{=}2$ & Gap \\
    \midrule
    42   & 70.17 & 71.29 & $+1.12$ \\
    123  & 70.05 & 71.46 & $+1.41$ \\
    456  & 69.95 & 71.17 & $+1.21$ \\
    777  & 69.84 & 71.02 & $+1.18$ \\
    2025 & 70.17 & 71.09 & $+0.92$ \\
    \midrule
    Mean$\pm$s.d. & $70.04\pm0.14$ & $71.21\pm0.17$ & $\mathbf{+1.17\pm0.18}$ \\
    \bottomrule
  \end{tabular}
\end{table}

\subsection{Reproducibility pointers}

The code repository at \url{https://github.com/libophd/sparse-moe-vision-rho} contains, under \texttt{phase2\_moe\_backbone/}:
\begin{itemize}
  \item \texttt{TIER\_C\_FINDINGS.md} --- canonical record with full per-seed numbers and source-JSON paths.
  \item \texttt{results\_c1\_rho/}, \texttt{results\_c2\_persample/}, \texttt{results\_c3\_k1/} --- 50 \texttt{results.json} files (30 + 15 + 5).
  \item \texttt{scripts/cluster/aggregate\_summary.py} --- aggregator with three schema parsers (\texttt{parse\_c1\_flat}, \texttt{parse\_c2\_nested}, \texttt{parse\_c3\_nested}).
\end{itemize}

\section{CIFAR-10 Development Configuration Sweep}
\label{sec:supp-cifar10-configs}

The CIFAR-10 development sweep over backbone width, expert hidden dimension, temperature schedule, and routing variants (referenced in Section~\ref{subsec:cifar10}). Multi-seed rows report mean$\pm$s.d.\ (final-epoch test accuracy); single-seed rows are exploratory references. $\Delta$FLOPs: negative $=$ reduction, positive $=$ increase vs.\ dense.

\begin{table}[ht]
  \centering
  \caption{CIFAR-10 configuration summary. \textsuperscript{$\ast$}Utility bias was found operationally negligible (Section~\ref{sec:supp-utility-ablation}).}
  \label{tab:supp-cifar10-configs}
  \footnotesize\setlength{\tabcolsep}{3pt}
  \begin{tabular}{@{}lcccc@{}}
    \toprule
    Config & Dense & MoE & Gap & $\Delta$FLOPs \\
    \midrule
    Wide 32/64 & 85.11 & 83.89 & $-1.22$ & $-$5.2\% \\
    Slim ($h{=}320$) & 83.39 & 83.28 & $-0.11$ & $-$8.3\% \\
    Slim ($h{=}304$) & 84.07 & 83.20 & $-0.87$ & $-$8.9\% \\
    Ultra-slim 16/32 & 82.51 & 81.15 & $-1.36$ & $-$13.5\% \\
    Sigmoid sched. & 83.38 & 83.41 & $+0.03$ & $-$8.9\% \\
    + utility bias\textsuperscript{$\ast$} & 83.34 & 84.26 & $+0.92$ & $-$9.4\% \\
    DW prototype & 78.03 & 77.40 & $-0.63$ & $-$22.0\% \\
    \midrule
    \multicolumn{5}{@{}l}{\emph{10-seed (Opt.\ DW, $w{=}0.72$):}} \\
    Mean$\pm$s.d. & 77.35$\pm$1.34 & 78.62$\pm$1.51 & $+1.28\pm1.26$ & $-$22.7\% \\
    \multicolumn{5}{@{}l}{\quad $p{=}.011$, $d{=}1.01$, CI $[+0.38, +2.18]$} \\
    \midrule
    \multicolumn{5}{@{}l}{\emph{10-seed (3blk+BN DW, $w{=}2.0$, opt.\ cfg):}} \\
    Mean$\pm$s.d. & 87.87$\pm$0.23 & 88.41$\pm$0.30 & $+0.54\pm0.16$ & $-$4.3\% \\
    \multicolumn{5}{@{}l}{\quad $p{<}10^{-5}$, $d{=}3.31$, CI $[+0.42, +0.65]$} \\
    \bottomrule
  \end{tabular}
\end{table}

\section{Utility Bias Ablation}
\label{sec:supp-utility-ablation}

The multi-dataset results in Section~\ref{sec:experiments} were obtained with an additive utility bias in the routing logits ($\lambda_u{=}0.087$, the value selected by the evolutionary search). To test whether this mechanism contributes to the observed gains, we ran a matched CIFAR-100 ablation with utility disabled ($\lambda_u{=}0$), holding all other hyperparameters fixed.

\begin{table}[ht]
  \centering
  \caption{Utility ablation. CIFAR-10: single-seed development reference. CIFAR-100: 5-seed mean$\pm$s.d.\ (transfer). \textsuperscript{$\dagger$}Paired test of utility effect on CIFAR-100: $p{=}0.87$.}
  \label{tab:supp-utility-ablation}
  \footnotesize\setlength{\tabcolsep}{3pt}
  \begin{tabular}{@{}llcccc@{}}
    \toprule
    Dataset & Configuration & Dense & MoE & Gap \\
    \midrule
    \multirow{3}{*}{C-10} & Std + utility & 83.34 & 84.26 & $+0.92$ \\
    & DW + no utility & 78.03 & 77.40 & $-0.63$ \\
    & DW + utility (opt.) & 77.35 & 78.62 & $+1.28$ \\
    \midrule
    \multirow{2}{*}{C-100\textsuperscript{$\dagger$}} & DW + no utility & $42.54\pm2.09$ & $45.63\pm2.72$ & $+3.09\pm1.63$ \\
    & DW + utility (opt.) & $41.38\pm2.10$ & $44.37\pm1.61$ & $+2.82\pm2.14$ \\
    \bottomrule
  \end{tabular}
\end{table}

On CIFAR-100 (transfer), disabling utility yields a gap statistically indistinguishable from the full recipe (paired $p{=}0.87$). Code inspection confirmed the cause: the additive utility bias was operationally negligible relative to learned router margins, not changing any top-$k$ selections in the tested models. The observed gains are therefore not attributable to the utility bias. The CIFAR-10 development rows are retained for completeness but should be interpreted in light of this finding.

\bibliographystyle{tmlr}
\bibliography{reference}

\end{document}